\documentclass{article}

\PassOptionsToPackage{numbers, compress}{natbib}

 \usepackage[main, final]{neurips_2026}

\usepackage{graphicx}
\usepackage{subcaption}
\usepackage{multirow}
\usepackage{floatrow}
\usepackage{tcolorbox}
\usepackage[utf8]{inputenc} 
\usepackage[T1]{fontenc}    
\usepackage{hyperref}       
\usepackage{url}            
\usepackage{booktabs}       
\usepackage{amsfonts}       
\usepackage{nicefrac}       
\usepackage{microtype}      
\usepackage[table,xcdraw]{xcolor}         
\usepackage{threeparttable}
\usepackage{color}
\usepackage{soul}
\usepackage{amsmath}
\usepackage{amssymb}
\usepackage{colortbl}
\usepackage{float}
\usepackage{wrapfig}
\usepackage{ragged2e}
\usepackage{adjustbox}
\usepackage{pifont}
\usepackage{algorithm}
\usepackage{algorithmic}
\usepackage{setspace}

\def\name{FaVChat}
\def\data{FaVChat-170K}
\def\fig{Fig. }
\def\tab{Tab. }
\def\sec{Sec. }
\usepackage{enumitem}
\setlist[itemize]{leftmargin=1em,itemsep=0pt,topsep=0pt}
\AtBeginDocument{%
  \setlength{\abovedisplayskip}{3pt plus 1pt minus 1pt}%
  \setlength{\belowdisplayskip}{3pt plus 1pt minus 1pt}%
  \setlength{\abovedisplayshortskip}{2pt plus 1pt minus 1pt}%
  \setlength{\belowdisplayshortskip}{2pt plus 1pt minus 1pt}%
}
\def\append{App. }

\usepackage[svgnames]{xcolor}

\usepackage{amsmath}
\usepackage{amssymb}
\usepackage{mathtools}
\usepackage{amsthm}
\title{\name: Hierarchical Prompt-Query Guided Facial Video Understanding with Data-Efficient GRPO}

\author{%
    Fufangchen Zhao$^{1}$,
    Songbai Tan$^{2}$,
    Xuerui Qiu$^{3,4}$,
    Linrui Xu$^{5}$,
    Wenhao Jiang$^{2}$,\\
    \textbf{Jinkai Zheng}$^{7}$,
    \textbf{Hehe Fan}$^{6}$,
    \textbf{Jian Gao}$^{1}$, 
    \textbf{Danfeng Yan}$^{1*}$,
    \textbf{Ming Li}$^{2*}$\\
    $^{1}$State Key Laboratory of Networking and Switching Technology, BUPT \\
    $^{2}$Guangdong Laboratory of Artificial Intelligence and Digital Economy (SZ) \\ $^{3}$Institute of automation, Chinese Academy of Sciences \\
    $^{4}$Zhongguancun Academy \quad $^{5}$Central South University \quad  $^{6}$Zhejiang University\\
    $^{7}$College of communication Engineering, Hangzhou Dianzi University \\
    {\tt\small zhaofufangchen.edu.cn} \\
}

\begin{document}

\maketitle

\def\thefootnote{*}\footnotetext{Corresponding author}

\begingroup
\setlength{\parskip}{0pt}%
\setlength{\belowcaptionskip}{5.5pt}%
\centering
\includegraphics[width=0.96\linewidth]{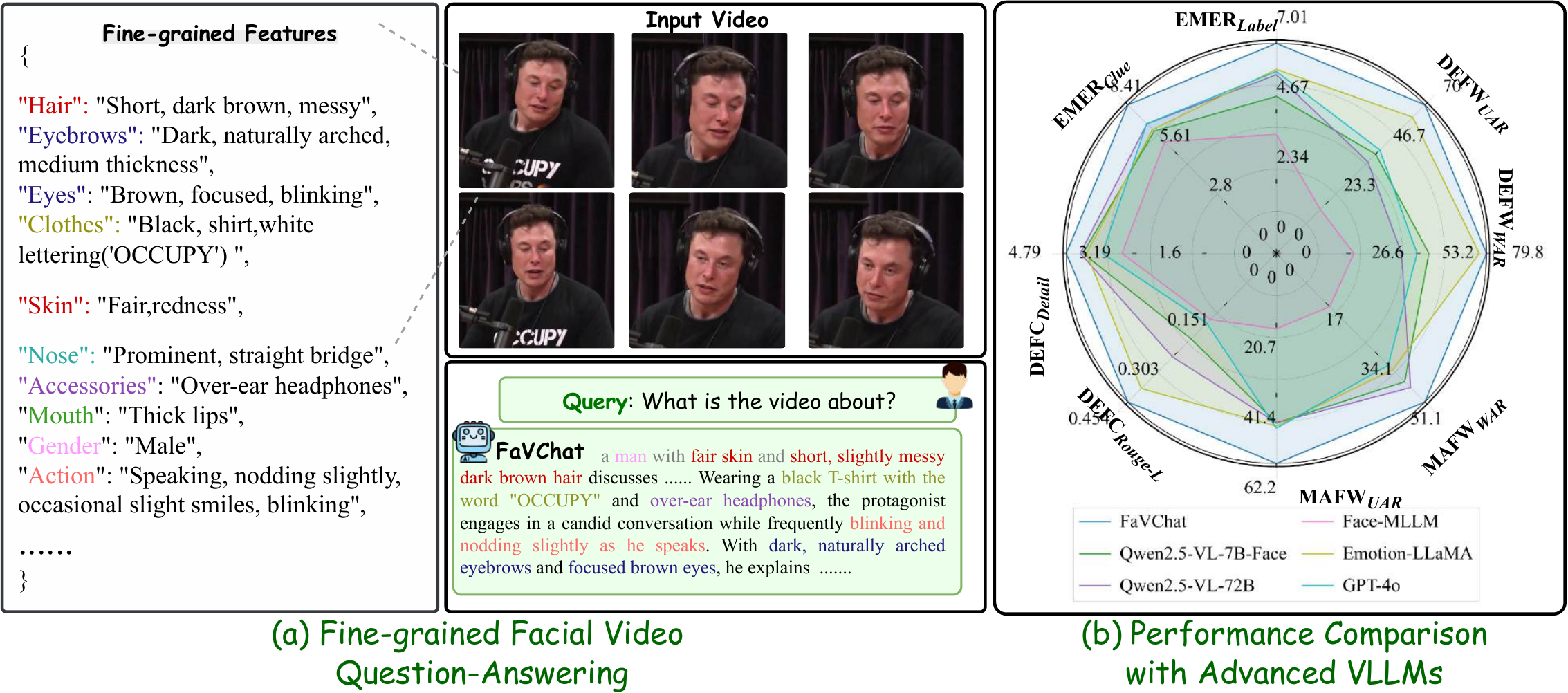}\par
\captionof{figure}{Overview and performance of \name{} for fine-grained facial video understanding. The left-side facial analysis is used internally and is not exposed in the end-to-end user experience.}
\par
\endgroup
\setlength{\parskip}{5.5pt}

\begin{abstract}
Existing video large language models (VLLMs) primarily leverage \textit{prompt-agnostic} visual encoders, which extract untargeted facial representations without awareness of the queried information, leading to the loss of task-critical cues.
To address this challenge, we propose \name{}, the \textbf{\textit{first}} VLLM designed for reasoning over subtle visual and dynamic facial cues. \name{} introduces a hierarchical, \textit{prompt-guided} visual feature extraction framework that emphasizes question-relevant information at three complementary levels.
These multi-level features are dynamically fused and injected into the LLM, enabling more accurate facial details reasoning.
To further improve learning efficiency under data scarcity, we propose Data-Efficient GRPO, a reinforcement learning strategy that iteratively identifies high-utility samples and maximizes the contribution of each instance via per-instance utility estimation, substantially enhancing performance gains under limited supervision.
We construct a large-scale benchmark dataset, \data{}, comprising approximately 60K high-quality facial videos and 170K question–answer pairs focusing on fine-grained facial details.
Extensive experiments, including zero-shot evaluations on four facial understanding tasks, demonstrate that \name{} consistently outperforms existing VLLMs.
\end{abstract}

\section{Introduction}
\label{sec:intro}

By integrating visual encoders \cite{li2023blip}, large language models, and projection layers \cite{alayrac2022flamingo, li2023blip}, multimodal large language models have significantly advanced vision–language understanding and enabled cross-modal reasoning \cite{zhu2023minigpt, alayrac2022flamingo, liu2023visual, chen2024internvl}. As video constitutes a primary medium for human perception and communication, video large language models (VLLMs) have emerged as a key frontier in multimodal research \cite{li2023videochat, lin2023video, li2024mvbench, fei2024video}.
Within this broader landscape, fine-grained facial video understanding plays a pivotal role, as it requires capturing subtle and localized facial cues—including skin texture, pupil characteristics, facial scars, and micro-level muscle movements—that are essential for real-world, human-centric intelligence \cite{lei2025ai, zhao2025humanomni, yang2025omniin}.

Despite this importance, existing VLLMs still face a fundamental challenge that limits their effectiveness in fine-grained facial understanding. Most current architectures rely on prompt-agnostic pure visual encoders, which extract representations without awareness of the queried information or task intent \cite{maaz2023video, luo2023valley, jin2024chat, li2024mvbench}. As a result, the learned visual features are coarse and untargeted, emphasizing generic semantics rather than question-relevant details. During hierarchical feature abstraction and evolution, subtle yet critical facial cues—such as fine textures, localized motions, and attribute-level variations—are consequently under-emphasized or discarded, hindering subsequent fine-grained comprehension in LLMs \cite{li2024mvbench,sun2024face}.

Moreover, although reinforcement learning has recently shown promise for improving multimodal alignment, existing label-free reward optimization strategies rely predominantly on coarse preference signals \cite{deng2025openvlthinker, liu2025visual, yang2025r1, zhang2025r1}. Such supervision is insufficient for guiding the modeling of dense and localized facial attributes \cite{guo2025deepseek}. Consequently, effective reward design would require detailed annotations. However, collecting explicit fine-grained supervision remains prohibitively labor-intensive and difficult to scale.

To address these challenges, we present \name{}, the first VLLM specifically designed for fine-grained facial video comprehension, as illustrated in Fig.~\ref{FaVChat}.
To overcome the limitation that query-relevant facial details are overlooked by prompt-agnostic visual encoders, \name{} introduces a hierarchical prompt-guided visual feature extraction framework that progressively captures question-relevant information at three complementary levels:
(1) low-level transformer features preserving fine textures and facial motions,
(2) mid-level learnable queries attending to discriminative facial regions, and
(3) high-level adaptive feature weighting for semantic alignment with language queries.
These enriched multi-level features are dynamically fused and injected into the LLM, enabling precise fine-grained understanding.

To further enable effective reinforcement learning under limited fine-grained supervision, we propose Data-Efficient (DE) GRPO, an iterative optimization strategy that estimates per-sample utility and dynamically schedules training to prioritize high-value samples.
With only 10K training samples, DE-GRPO achieves significantly higher performance than the vanilla GRPO counterpart trained with substantially more data (170K as shown in \tab\ref{rl_ablation}).
To support systematic training and evaluation, we devise a meticulous multi-step pipeline to construct a large-scale, high-quality cross-modal facial video dataset, \data{}, comprising over 60K carefully curated videos with rich captions and 170K question–answer (QA) pairs spanning 103 facial attributes, including appearance, actions, and emotions. Extensive experiments demonstrate that \name{} achieves state-of-the-art performance on our benchmark as well as four zero-shot evaluation tasks. Notably, \name{} (7B) outperforms Qwen2.5-VL-72B by 26.87 UAR and 32.27 WAR on the DFEW benchmark, despite using a substantially smaller model.

In summary, our main contributions are as follows:
\begin{itemize}
\item We present \name{}, the first VLLM explicitly designed for fine-grained facial video comprehension, addressing a long-standing challenge in detailed facial analysis for multimodal understanding.

\item We propose a hierarchical prompt-guided visual feature extraction framework that enables selective extraction of task-relevant cues while preserving subtle facial details for fine-grained reasoning.

\item We introduce DE-GRPO, a data-efficient reinforcement learning algorithm that replaces coarse preference supervision with fine-grained optimization via per-sample utility estimation and dynamic lifecycle scheduling, substantially improving data efficiency under limited supervision.

\item We contribute a large-scale, high-quality cross-modal facial video dataset, \data{}, comprising over 60K videos and 170K question–answer pairs covering 103 facial attributes across diverse appearances, actions, and emotions.
\end{itemize}
\setlength{\parskip}{4pt}
\section{FaVChat}

\begin{figure*}[t]
   \centering
   \includegraphics[width=0.86\linewidth]{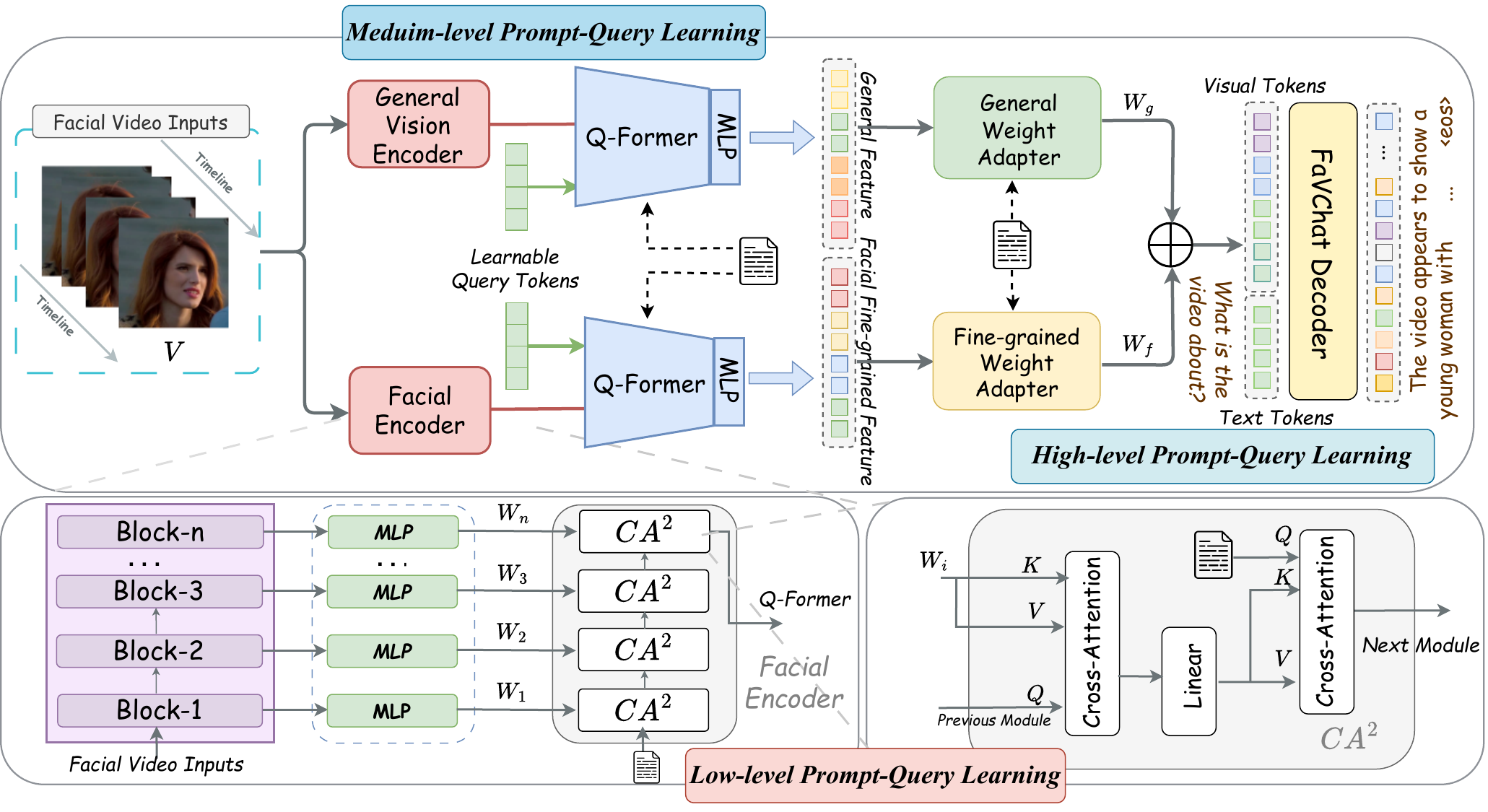}
    \vspace{-2mm}
    \caption{Overview of \name. We augment the general visual encoder with a facial encoder \cite{narayan2024facexformer} and use hierarchical prompt-query learning: low-level cross-layer cue aggregation, mid-level learnable queries, and high-level query-conditioned weighting.}
    \label{FaVChat}

\end{figure*}

In this section, we present \name{}, a MLLM framework designed for fine-grained facial video understanding. 
As illustrated in \fig\ref{FaVChat}, \name{} is built upon two tightly coupled components: a hierarchical prompt-query visual encoding that progressively preserves the fine-grained facial cues during visual encoding, and a data-efficient reinforcement learning strategy that enables effective fine-grained alignment under limited data.



\subsection{Hierarchical Prompt-Query Visual Encoding}
\paragraph{Motivation.}
Existing video MLLMs usually encode visual inputs into a single set of text-agnostic embeddings, where textual prompts are only introduced at the final fusion stage. This design assumes that all task-relevant facial information has already been preserved during visual encoding. However, this design is insufficient in capturing the detailed facial features that are distributed across different spatial areas and network depths. We argue that fine-grained facial understanding requires textual prompts to guide visual encoding throughout the entire hierarchy, rather than being applied only after visual feature. Motivated by this, we decompose visual encoding into three prompt-query stages:
(1) \textbf{low-level prompt-query learning} for retaining fine-grained facial cues, (2) \textbf{mid-level prompt-query learning} for extracting discriminative facial regions, and (3) \textbf{high-level prompt-query learning} for semantically adaptive fusion of visual features.

\paragraph{Low-level Prompt-Query Learning.}
As shown in Fig. \ref{FaVChat}, we utilize a dedicated facial encoder and a general vision encoder to extract facial fine-grained and general features, respectively. Within the facial encoder, we integrate a low-level prompt-query learning mechanism that enables textual queries to attend to intermediate spatiotemporal features, steering the aggregation of relevant details across Transformer layers. Specifically, the facial encoder processes an input video $\mathbf{V} \in \mathbb{R}^{T \times C \times H \times W}$ through Transformer blocks, where the $i$-th block outputs features $f_i(\mathbf{V}) \in \mathbb{R}^{T \times P_i \times D_i}$, capturing information at different granularity. To facilitate cross-layer interaction, these features are projected into a shared embedding space:

\begin{equation}
\mathbf{W}_i = f_i(\mathbf{V}) W_i , \quad \mathbf{W}_i \in \mathbb{R}^{T \times P_i \times D}.
\end{equation}

Instead of relying solely on the final layer, we adopt a progressive prompt-query aggregation strategy based on a sequence of $CA^2$ modules.
In the first module, the textual embedding $\mathbf{X}$ serves as the query to attend to the projected features $\mathbf{W}_1$:

\begin{equation}
Q_1 = \mathbf{X} W_Q, \quad K_1 = \mathbf{W}_1 W_K, \quad V_1 = \mathbf{W}_1 W_V.
\end{equation}
For subsequent layers ($i = 2, \ldots, N$), we use the output of the preceding $CA^2$ module as the query to attend to the current visual features:

\begin{equation}
Q_i = E_{i-1}, \quad K_i = \mathbf{W}_i W_K, \quad V_i = \mathbf{W}_i W_V,
\end{equation}
\begin{equation}
E_i = \text{softmax}\left(\frac{Q_i K_i^\top}{\sqrt{d}}\right) V_i .
\end{equation}
The final output $E_N$ is taken as the facial fine-grained feature. Compared with directly adopting the final-layer feature, this approach better preserves subtle facial details from shallow layers while maintaining semantic alignment with the prompt-query.

\paragraph{Medium-level Prompt-Query Learning.}
Although low-level prompt-query learning preserves fine-grained cues, the resulting features remain dense and lack explicit semantic abstraction. Given the complementary inductive biases between the general vision encoder and the facial encoder, we introduce Q-Formers to extract discriminative mid-level representations. Each Q-Former uses learnable query tokens, guided by textual prompts, to selectively aggregate query-relevant visual information via cross-attention. This design converts dense visual features into compact, query-aligned tokens with distinct semantic roles, facilitating effective prompt-query guiding.

\paragraph{High-level Prompt-Query Learning.}
\label{high_level}
High-level prompt-query learning is designed to model the query-dependent importance of visual evidence derived from different encoders. Treating the general and facial streams as equally informative risks introducing irrelevant or redundant evidence. To address this issue, we introduce two text-conditioned weight adapters that estimate the contribution of the general and facial fine-grained features. Specifically, given visual tokens $\mathbf{V}_g, \mathbf{V}_f \in \mathbb{R}^{L \times d}$ and the textual representation $\mathbf{T} \in \mathbb{R}^{L \times d}$, each adapter evaluates the alignment between its visual features and the input query, producing a scalar confidence score:

\begin{equation}
\small
s_g = \mathcal{A}_g(\mathbf{V}_g, \mathbf{T}), \quad
s_f = \mathcal{A}_f(\mathbf{V}_f, \mathbf{T}),
\end{equation}
where $\mathcal{A}_g(\cdot)$ and $\mathcal{A}_f(\cdot)$ denote the general and fine-grained weight adapters, respectively. The resulting scores are normalized to obtain importance weights:

\begin{equation}
\small
W_g=\frac{e^{\mathbf{s}_g}}{e^{\mathbf{s}_g}+e^{\mathbf{s}_f}}, \quad W_f=\frac{e^{\mathbf{s}_f}}{e^{\mathbf{s}_g}+e^{\mathbf{s}_f}}.
\end{equation}
These weights dynamically regulate the contribution of each visual stream when conditioning the LLM, enabling adaptive emphasis on global structures or fine-grained details based on the query’s semantic intent.

\subsection{Data Efficiency-GRPO}
\label{method:rl}
\begin{wrapfigure}[11]{r}{7.4cm}
  \caption{Concept diagrams of GRPO and DE-GRPO.}
  \vspace{-4mm}
  \begin{adjustbox}{max width=\linewidth}
    \includegraphics[width=\textwidth]{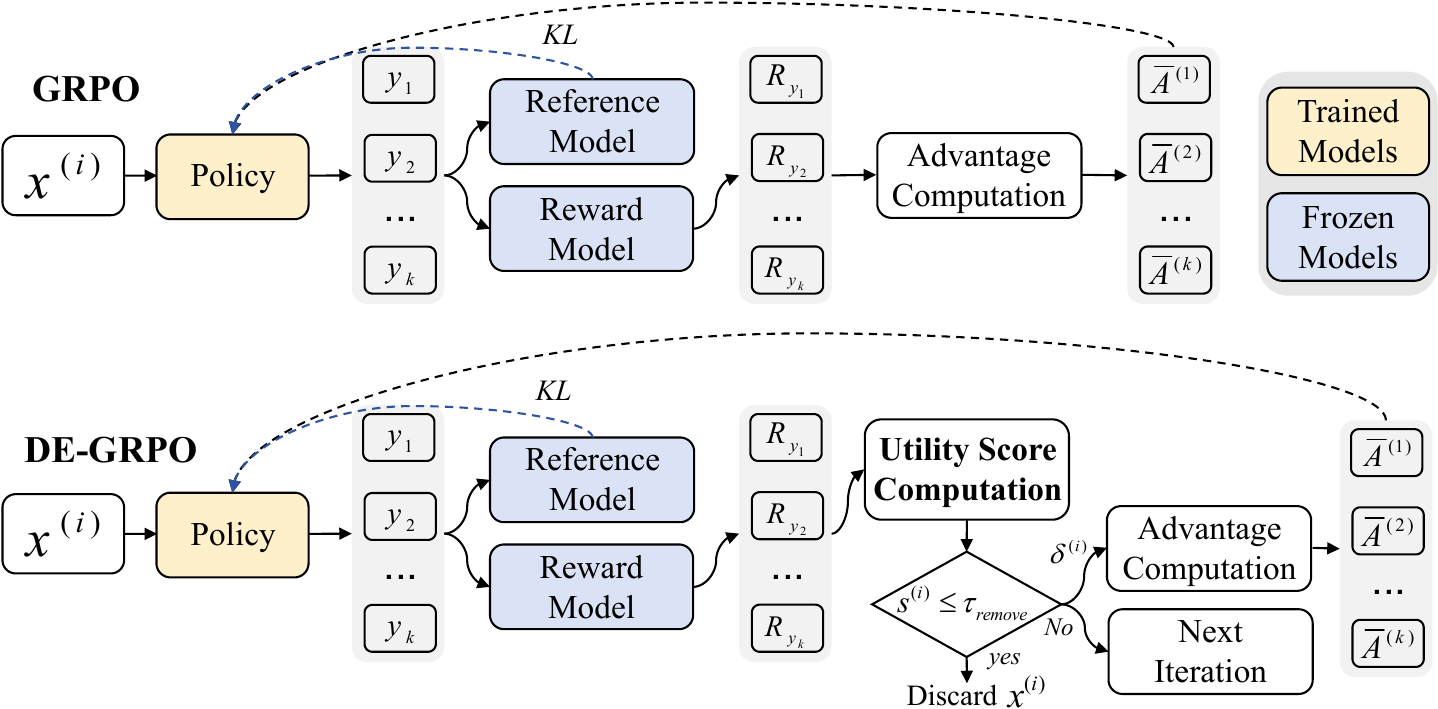}
  \end{adjustbox}
  \vspace{-4mm}
  \label{de-grpo}
\end{wrapfigure}


We propose Data-Efficient GRPO (DE-GRPO), a reinforcement learning framework designed for fine-grained facial video understanding under limited high-quality supervision.
Standard GRPO~\cite{guo2025deepseek} and variants such as DrGRPO~\cite{liu2025understanding} and DAPO~\cite{yu2025dapo} use group statistics mainly to normalize response-level rewards and treat sampled data as static. This is suboptimal for fine-grained facial reasoning, where some videos provide subtle, high-value supervision while others are easy or ambiguous. DE-GRPO departs from GRPO by (i) replacing coarse preference or format rewards with a facial-specific reward over attributes, emotions, and actions, and (ii) reusing response-group statistics as a \emph{data utility probe} to decide whether the input video itself deserves further optimization. As shown in \fig\ref{de-grpo}, GRPO-style grouping is thus extended from comparing responses to managing sample lifecycles.

\paragraph{Facial-Specific Fine-Grained Reward.}
Preference-only rewards are too coarse to supervise dense facial semantics, since two responses can have similar overall preference scores while differing in key details such as eye motion, facial hair, or micro-expressions.
We therefore define a structured reward that explicitly evaluates generated responses against fine-grained facial annotations.
Given a generated response $y_i$, the reward is
\begin{equation}
\small
\boldsymbol{R}_{\text {}}(y_i)=\sum_{j\in\{attr,emo,act\}} {\alpha_j} \cdot \operatorname{Sim}\left(y_i,y^*_{i,j}\right),
\end{equation}
where $y_{i,j}^*$ denotes the ground-truth annotation of facial \emph{attributes}, \emph{emotions}, and \emph{actions}, respectively.
$\operatorname{Sim}(\cdot)$ measures semantic similarity, and $\alpha_j$ denotes the normalized weight for each facial semantic category.
This reward provides dense, task-specific supervision and makes the reward differences between candidate responses interpretable as differences in fine-grained facial correctness.

\paragraph{Utility-Aware Recurrent Data Mechanism.}
The central contribution of DE-GRPO is to model data efficiency at the sample level.
For each facial video sample $x^{(i)}$, the current policy $\pi_\theta$ generates $K~(K\geq 4)$ candidate responses.
From these responses, we construct a set of preference pairs $\mathcal{P}$ and estimate the optimization utility of the input sample along two complementary dimensions.
The first dimension is reward separability, which measures whether the fine-grained reward can clearly distinguish winning and losing responses:
\begin{equation}
\label{eq_8}
\small
\hat R^{(i)}=\left\{\prod_{\left(y_w, y_l\right) \in \mathcal{P}}\left|\boldsymbol{R}_{\text {}}\left(y_w^{(i)}\right)-\boldsymbol{R}_{\text {}}\left(y_l^{(i)}\right)\right|\right\}^{\frac{1}{|\mathcal{P}|}},
\end{equation}
where $y_w$ and $y_l$ denote the winning and losing responses in each pair, respectively.
A larger $\hat R^{(i)}$ indicates that the sample yields a clearer supervision signal.
The second dimension is gradient sensitivity, which measures whether different responses to the same video induce substantially different policy updates:
\begin{equation}
\label{eq_9}
\small
\hat g^{(i)} = \left\{\prod_{\left(y_w, y_l\right) \in \mathcal{P}}\left\|g(y^{(i)}_w)-g(y^{(i)}_l)\right\|_2\right\}^{\frac{1}{|\mathcal{P}|}},
\end{equation}
where $g(y^{(i)}) = \nabla_\theta \log \pi_\theta(y^{(i)} \mid x^{(i)})$.
A larger $\hat g^{(i)}$ suggests that the policy is uncertain about how to model the sample, and that the sample still has optimization potential.
Combining these two terms, the sample utility is defined as
\begin{equation}
\label{U_i}
U^{(i)}=\hat R^{(i)}\cdot\hat g^{(i)}.
\end{equation}
Thus, DE-GRPO prioritizes samples that are both semantically separable and optimization-sensitive, rather than simply replaying all available data.

\paragraph{Dynamic Sample Lifecycle and Decay Rationale.}
To capture the long-term usefulness of each sample, DE-GRPO maintains a recurrent state variable $s^{(i)}$ and updates it across training iterations by exponential smoothing:
\begin{equation}
\label{s^i}
s^{(i+1)}=\lambda \cdot s^{(i)}+(1-\lambda) \cdot \mathbb{I}\left[U^{(i)}>\tau\right],
\end{equation}
where $i$ denotes the iteration step, $\lambda$ controls temporal smoothing, and $\tau$ is a batch-wise utility threshold.
This recurrent state prevents the training schedule from being determined by a single noisy utility estimate.
Based on $s^{(i)}$, each sample is assigned one of three lifecycle states:
\begin{equation}
\label{delta}
\delta^{(i)}=
\begin{cases}
\text {removed}, & s^{(i)} \leq \tau_{\text {remove}},\\
\delta, & s^{(i)} \geq \tau_{\text {keep}},\\
1, & \text {otherwise,}
\end{cases}
\end{equation}
where $\tau_{\text {remove}}$ and $\tau_{\text {keep}}$ are lower and upper utility thresholds, respectively, and $\delta \in (0,1)$ is the decay factor.
The decay factor is deliberately set below one for samples whose recurrent utility exceeds $\tau_{\text {keep}}$.
Although these samples are informative, assigning them a weight $\geq 1$ would allow their accumulated utility to grow monotonically, causing them to dominate later batches and increasing the risk of overfitting to a small subset of high-reward videos.
By contrast, $\delta<1$ implements a cooling-down strategy: once a sample has been discovered and exploited, its influence is reduced to reflect the diminishing marginal return of repeated training.
Samples in the intermediate range keep full weight, allowing newly informative examples to accumulate evidence before being retained or removed.
This design forms a continuous data lifecycle of discovery, exploitation, cooling, and removal, which is the key mechanism enabling DE-GRPO to improve data efficiency rather than merely filtering low-score samples.

\paragraph{Sample-Level Advantage.}
Another important distinction from standard GRPO lies in the granularity of the advantage function.
In standard GRPO, the advantage is computed for each generated response and serves as a response-level alignment signal relative to the group baseline.
DE-GRPO instead optimizes the utility of the input video sample.
Therefore, after the preference pairs $\mathcal{P}$ are constructed for the same input $x^{(i)}$, we compute an average reward difference and use it to form a sample-level advantage:
\begin{equation}
\label{A^i}
\small
\begin{split}
\Delta R^{(i)} &= \frac{1}{|\mathcal{P}|}\sum_{\left(y_w, y_l\right) \in \mathcal{P}}\left(\boldsymbol{R}_{\text {}}(y_w^{(i)})-\boldsymbol{R}_{\text {}}(y_l^{(i)})\right),\\
\overline{\boldsymbol{A}}^{(i)} &= \delta^{(i)} \cdot \left[\Delta R^{(i)} - \boldsymbol{b}_\phi(x^{(i)})\right],
\end{split}
\end{equation}
where $\boldsymbol{b}_\phi(x^{(i)})$ is a learned value model that estimates the expected return of the input sample and reduces variance.
Here, $\overline{\boldsymbol{A}}^{(i)}$ is shared by all response pairs generated from the same video and functions as a \emph{sample-level importance weight}, not as an individual response-level score.
It scales the policy update according to how much useful supervision the input video currently provides: samples with clearly separable rewards and high optimization sensitivity contribute stronger gradients, while exhausted or low-utility samples are down-weighted or removed.
The resulting objective follows a GRPO-style clipped update with KL regularization, but its optimization mass is adaptively redistributed across data according to sample utility.
By coupling fine-grained facial reward with recurrent sample lifecycle control, DE-GRPO contributes a data-centric reinforcement learning strategy for MLLMs, enabling effective fine-grained alignment with substantially fewer training samples.
The complete objective and algorithmic procedure are provided in \append\ref{appendix:rl}.

\subsection{Progressive Pre-training Paradigm}
\label{train stratege}
\begin{wrapfigure}[11]{r}{7.6cm}
\vspace{-2mm}
  \caption{An illustration of our proposed pre-training paradigm.}
  \vspace{-4mm}
  \begin{adjustbox}{max width=\linewidth}
    \includegraphics[width=\textwidth]{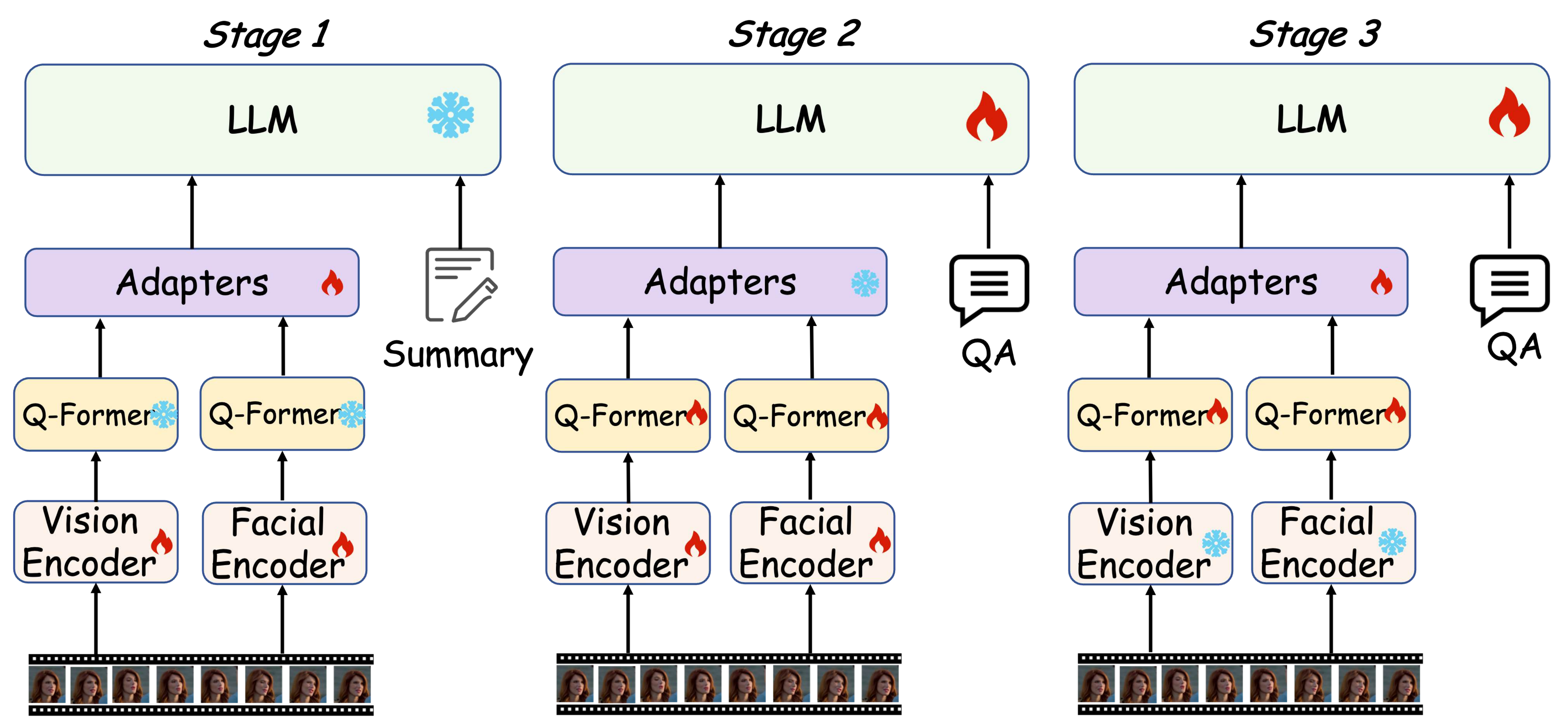}
  \end{adjustbox}
  \vspace{-4mm}
  \label{train}
\end{wrapfigure}


As illustrated in Fig. \ref{train}, we adopt progressive pre-training for hierarchical prompt-query encoding. First, visual encoders and adapters are trained on 60K video-summary pairs while Q-Formers and LLM are frozen, preserving low-level cues. Second, we use 170K video QA pairs and unfreeze Q-Formers while keeping adapters fixed, enabling prompt-guided queries to aggregate discriminative facial regions. Finally, 110K emotion-intensive QA pairs train the adapters, Q-Formers, and LLM with frozen visual encoders for high-level semantic adaptation. This progression learns low-level details, mid-level regions, and high-level relevance in order.

\section{Experiments}
\label{sec:experiments}

\subsection{FaVChat-170K Dataset}
\label{data_collec}
\begin{figure*}[!htb]
  \centering
  \begin{minipage}{0.83\textwidth}
    \centering
    \begin{adjustbox}{valign=t,max width=\linewidth}
    \begin{tabular}{@{}lccccccc@{}}
    \toprule
      \multirow{2}{*}{\textbf{Dataset}}   & \multirow{2}{*}{\textbf{Video Resolution}} & \textbf{VQA} & \textbf{VQA pairs} & \multirow{2}{*}{\textbf{Video}} &\textbf{Caption Tokens} & \textbf{Attributes} & \textbf{Attributes Number} \\
       & & \textbf{Availability} &\textbf{(Avg.)} & & \textbf{(Avg.)} &\textbf{(Avg. per Caption)} & \textbf{(total)} \\
       \midrule
       HMDB51 \cite{wishart2022hmdb} & \multirow{2}{*}{\textbf{Low}} & \textcolor{red}{\ding{55}} & \textcolor{red}{\ding{55}} & 6766 & 13.1 & 2.3 & 23 \\
       Youtube-face \cite{wolf2011face} & & \textcolor{red}{\ding{55}} & \textcolor{red}{\ding{55}} & 3425 & \textcolor{red}{\ding{55}} & \textcolor{red}{\ding{55}} & 11 \\
       \midrule
       FERV39K \cite{wang2022ferv39k} & \multirow{3}{*}{\textbf{High}} & \textcolor{red}{\ding{55}} & \textcolor{red}{\ding{55}} & 39546 & 11.3 & 1.3 & 7 \\
       CelebV-HQ \cite{zhu2022celebv} &  & \textcolor{red}{\ding{55}} & \textcolor{red}{\ding{55}} & 35666 & \textcolor{red}{\ding{55}} & \textcolor{red}{\ding{55}} & 83 \\
        \rowcolor{gray!15}
        \textbf{FaVChat-170K} & & \textcolor{green}{\ding{51}}(\textbf{170K}) & \textcolor{green}{\ding{51}}(2.8) & 61007 & 100.7 & 23.4 & 103 \\
        \toprule

    \end{tabular}
    \end{adjustbox}
  \end{minipage}
\hfill
  \begin{minipage}{0.15\textwidth}
    \centering
    \adjustbox{valign=t}{\includegraphics[width=\linewidth]{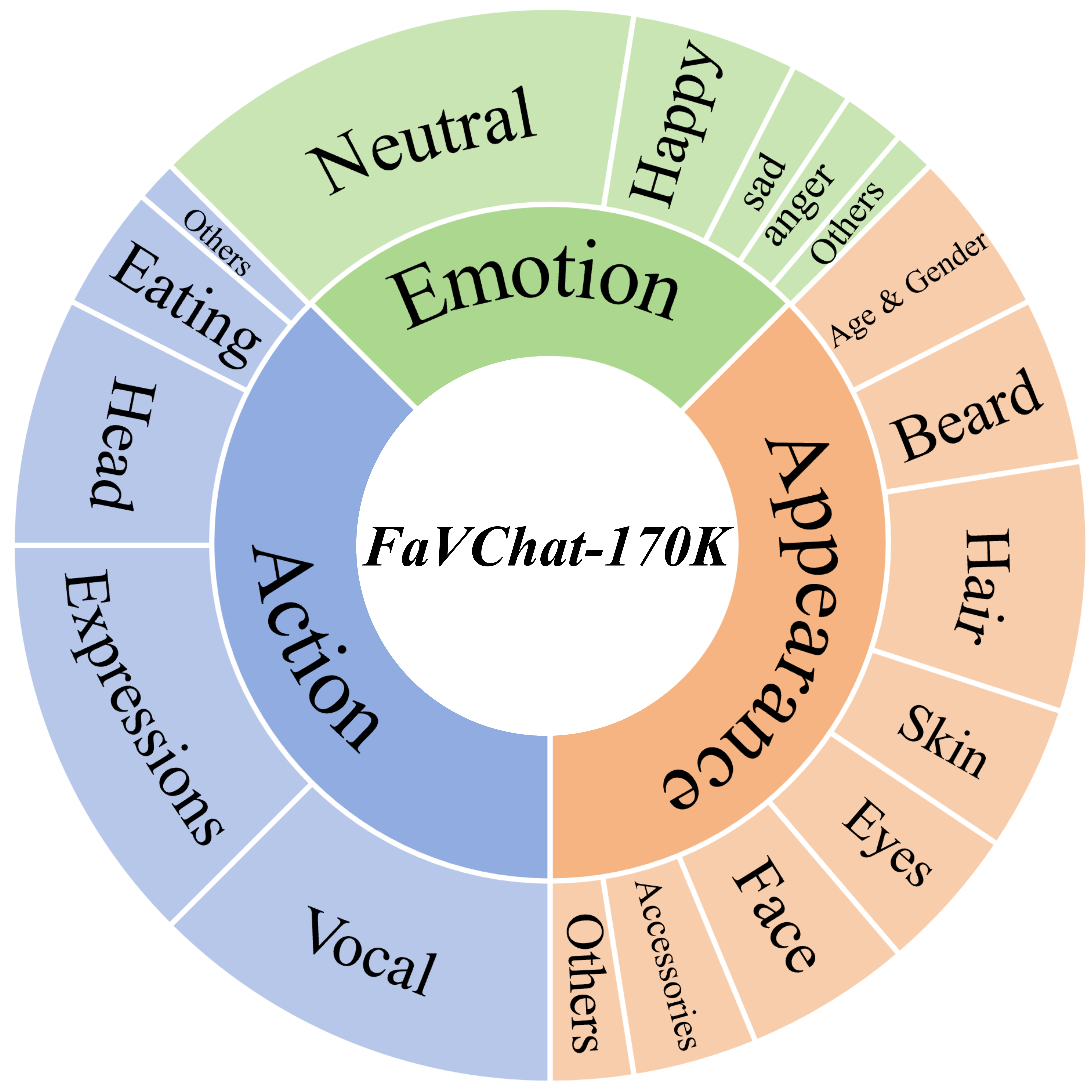}}
  \end{minipage}
  \vspace{-3mm}
  \caption{Dataset comparison. \data{} contains the most videos, the richest facial attributes, and the first face-video VQA training data.}
  \label{fig:dataset_comparison}
\end{figure*}

To support fine-grained facial video understanding, we construct a large-scale dataset by curating and annotating videos from CelebV-HQ \citep{zhu2022celebv}, FERV39K \citep{wang2022ferv39k}, HMDB51 \citep{kuehne2011hmdb}, and Youtube Faces \citep{DBLP:conf/cvpr/WolfHM11}, covering diverse identities, expressions, and actions. CelebV-HQ contains 35,666 clips across 15,653 subjects with 83 manually labeled facial attributes, while 25,341 additional videos from the other datasets provide varied facial motions. To compensate for missing textual annotations, we train a feature extraction module with three extractors to unify 103 fine-grained features per video, followed by GPT-4o \citep{hurst2024gpt} to generate structured video summaries and QA pairs. This results in 60K high-quality video-summary pairs and 170K QA pairs, forming a comprehensive benchmark for fine-grained facial video understanding. As shown in \fig\ref{fig:dataset_comparison}, further dataset details are provided in \append\ref{dataset}.

\subsection{Quantitative Comparison}
\label{Quantitative Comparison}
To demonstrate \name's advantages, we construct a 2,000-sample held-out test set from our internal dataset. These samples are excluded from training and annotated with detailed fine-grained labels and high-quality textual descriptions. We evaluate \name~ and several baseline models on this test set using the VideoChatGPT score, extending the standard 5-point scale to 10 for more discriminative assessment. The corresponding experimental results are highlighted with $*$.
\begin{table*}[h]
    \centering
    \small
    \setlength{\tabcolsep}{1.8pt} 
    \caption{\small
Performance comparison on DFEC \citep{zhao2025facial} and our held-out test set. $*$ denotes VideoChatGPT Scores on our test set; $\clubsuit$ denotes visual-speech models; ``-Face'' denotes variants fine-tuned on \data{}. Complete results are in \append\ref{appendix:expression_complete}. 
    }
    \label{expression}
  \begin{adjustbox}{max width=0.85\linewidth} 
    \begin{tabular}{lcccccccccc}
        \hline
        \multirow{2}{*}{\textbf{Method}} & \textbf{LLM}  & \multicolumn{5}{c}{\textbf{VideoChatGPT Scores}} & \multicolumn{2}{c}{\textbf{N-gram Based}} &\multicolumn{1}{c}{\textbf{Event Based}} \\
        \cmidrule(lr){3-7} \cmidrule(lr){8-9}\cmidrule(lr){10-10}
        & \textbf{Size} &{\textbf{Test(2000)$^*$}} &\textbf{Correctness} & \textbf{Detail} & \textbf{Context} & \textbf{Temporal} & \textbf{CIDEr} & \textbf{Rouge-L} & \textbf{AutoDQ} \\
        \hline
        \rowcolor{gray!20}
        \multicolumn{10}{c}{\textbf{\textit{General VLLM}}} \\ \hline
        ShareGPT4Video \citep{chen2024sharegpt4video} &7B & 7.33& 4.24 & 4.13 & 4.35 &  4.09 & 0.192 & 0.205 & 0.394  \\
        VideoLLaMA2 \citep{cheng2024videollama} &7B &7.41 & 4.17 & 4.02 & 4.47 & 3.93 & 0.253 & 0.266 & 0.344  \\
        PLLaVA \citep{xu2024pllava} &7B & 7.47 & 4.21 & 4.15 & 4.37 & 4.08 & 0.268 & 0.250 & 0.393  \\
        Qwen2.5-VL-7B \citep{Qwen2.5-VL} &7B  & 7.84 & 4.49 & 4.36 & 4.43 & 4.21 & 0.261 & 0.254 & 0.417  \\
        Qwen2.5-VL-72B \citep{Qwen2.5-VL} &72B & 8.47 &4.61 & 4.47 & 4.59 & 4.78 & 0.281 & 0.317 & 0.453 \\
        Qwen3-VL-8B \cite{bai2025qwen3vltechnicalreport} & 8B & 7.99 & 4.49 & 4.37 & 4.41 & 4.33 & 0.262 & 0.257 & 0.421 \\
        Qwen3-VL-30B-A3B \cite{bai2025qwen3vltechnicalreport} & 30B & 8.44 & 4.63 & 4.51 & 4.57 & 4.81 & 0.304 & 0.331 & 0.461 \\
        Claude4-Sonnet \citep{Claude2025} & - & 7.84 &4.55 &4.64 &4.77 &4.84 &0.312 &0.342 &0.472\\
        GPT-4o \citep{openai2024gpt4o} & - & 7.67 &4.22 &3.97 &4.48 & 3.90 & 0.264 &0.213 &0.432 \\
        \hline
         \rowcolor{gray!20}
        \multicolumn{10}{c}{\textbf{\textit{Human-Centric VLLM}}} \\ 
        \hline
        FaceTrack-MM \cite{zhao2025facial} & 7B & 7.47 & 4.42 & 4.30 & 4.60 & 4.26 & 0.418 & 0.473 & 0.483 \\
        Qwen2.5-VL-7B-Face \cite{Qwen2.5-VL} & 7B & 8.64 & 4.51 & 4.41 & 4.45 & 4.22 & 0.274 & 0.261 & 0.463 \\
        Qwen3-VL-8B-Face \cite{bai2025qwen3vltechnicalreport} & 8B & 8.87 & 4.58 & 4.46 & 4.47 & 4.51 & 0.297 & 0.312 & 0.429 \\
        Emotion-LLaMA$^\clubsuit$ \cite{cheng2025emotion} & 7B & 7.42 & 4.17 & 4.33 & 4.54 & 4.19 & 0.294 & 0.413 & 0.372\\
        HumanOmni$^\clubsuit$ \cite{zhao2025humanomni} & 7B & 7.51 & 4.58 & 4.41 & 4.70 & 4.41 & 0.412 & 0.468 & \textbf{0.523} \\
        Omni-Emotion$^\clubsuit$ \cite{yang2025omniin} & 7B & 7.53 & 4.57 & 4.38 & 4.73 & 4.24 & 0.407 & 0.461 & 0.504 \\
        \cmidrule(lr){1-10} 
        \textbf{FaVChat$_{\mathrm{Qwen2.5}}$} & 7B & 8.89 &4.84 &4.79 & 4.69 &4.78 &0.443 &0.454 &0.487 \\
        \textbf{FaVChat$_{\mathrm{Qwen3}}$} & 8B & \textbf{9.01} & \textbf{4.87} & \textbf{4.82} & \textbf{4.74} & \textbf{4.81} &\textbf{0.447} & \textbf{0.469} & \textbf{0.523} \\
        \hline
    \end{tabular}\end{adjustbox}
    
\end{table*}

\begin{figure*}[t]
   \centering
   \includegraphics[width=0.83\linewidth]{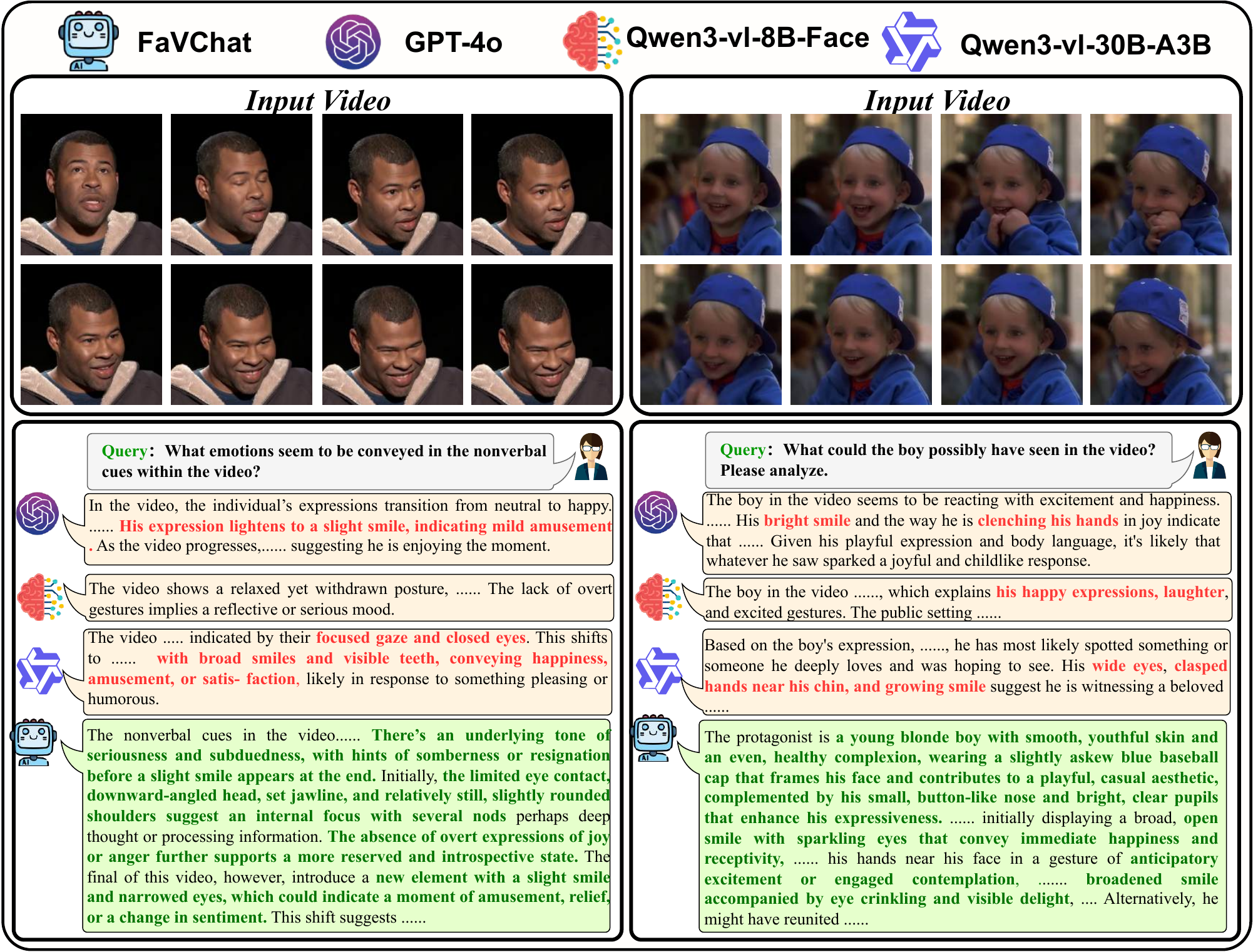}
    \caption{Qualitative comparisons show FaVChat outperforms leading VLLMs in facial video QA. Given identical prompts, FaVChat captures detailed facial features, expressions, and movements, whereas others focus on general appearance and backgrounds, demonstrating its superior fine-grained facial video understanding.}
    \label{table:qualitative}
\end{figure*}

The results, presented in Table \ref{expression}, show that \name~ consistently outperforms existing methods across nearly all metrics. Even in N-gram-based evaluations, where general models often struggle, \name~ remains highly competitive. We attribute this to its ability to extract fine-grained facial features and subtle motion cues, leading to more contextually accurate and emotionally aligned predictions. These results highlight \name’s strong capability in fine-grained facial video understanding, which is essential for precise emotion analysis.

\subsection{Qualitative Comparison}

\fig\ref{table:qualitative} compares the facial video QA capabilities of FaVChat with several leading VMLLMs. FaVChat generates the most detailed and expressive descriptions, capturing fine-grained facial attributes, skin texture, accessories, and subtle expression changes (e.g., “downward-angled head, rounded shoulders, nods, slight smile, narrowed eyes”). In contrast, models such as Qwen3-vl-8B-Face \cite{bai2025qwen3vltechnicalreport}, GPT-4o \cite{hurst2024gpt}, and Qwen3-vl-30B-A3B \cite{bai2025qwen3vltechnicalreport} focus mainly on general appearance, clothing, and background, missing many facial details and dynamic cues. These results demonstrate \name’s strong fine-grained facial understanding, making it well suited for tasks such as emotion recognition and human–computer interaction.

\subsection{Zero-shot Generalizations}

\paragraph{Textual Facial Analysis.}
Facial expressions reflect internal emotional states such as happiness, sadness, or anger through subtle movements like smiling, frowning, or smirking. Yet a single expression can map to multiple emotions, making facial emotion analysis inherently ambiguous. Accurate recognition therefore requires fine-grained feature extraction and precise motion modeling. Given FaVChat’s strength in capturing subtle facial dynamics, we evaluate its effectiveness on this task using the recently released DFEC dataset \citep{zhao2025facial} and its recommended metrics. As shown in \tab\ref{expression}, we conducted supplementary evaluations of FaVChat using Qwen3 \cite{yang2025qwen3} as the backbone model. Notably, to ensure a fair comparison, all results presented in other generalizations are derived from FaVChat evaluations with Qwen2.5-base \cite{qwen2.5} serving as the backbone.

\paragraph{Emotion Classification.}

\begin{wraptable}{r}{8.0cm}
    \vspace{-2mm}
    \caption{Comparison with existing methods on two large in-the-wild emotion classification datasets.}
    \label{emotion}
    \begin{adjustbox}{max width=0.98\linewidth}
         \begin{tabular}{lccccc}
    \\
    \hline
       Dataset  &  & \multicolumn{2}{c}{DFEW  \citep{jiang2020dfew}} & \multicolumn{2}{c}{MAFW\citep{liu2022mafw}} \\ \hline
        Methods & modality & UAR & WAR &  UAR & WAR  \\  \hline
TMEP \citep{zhang2023transformer}& AV & 57.16 & 68.85 &37.17 & 51.15  \\
    HiCMAE   \citep{sun2024hicmae}   & AV    &   63.76    &  75.01   & 42.65   &  56.17      \\ 
    Emotion-LLaMA \citep{cheng2025emotion}& AV & 64.21 & 77.06 & - & -  \\
    MMA-DFER \citep{chumachenko2024mma} & AV& 66.01 & 77.51 & 44.11 & 58.52 \\
    \midrule
DFER-CLIP  \citep{zhao2023prompting}   & V  & 59.61  & 71.25       & 38.89   & 52.55    \\
SVFAP \citep{sun2024svfap}& V & 62.83 & 74.27 & 41.19 & 54.28   \\

    MAE-DFER \citep{sun2023mae}&V & 63.41 & 74.43 & 41.62 & 54.31 \\
        Finecliper \citep{chen2024finecliper}& V & 65.98 & 76.21 & 45.01 & 56.91 \\
        Qwen2.5-VL-7B-Face \citep{Qwen2.5-VL} &V &46.78 &57.93 &44.13 &50.17 \\
    Qwen2.5-VL-72B \citep{Qwen2.5-VL} &V &43.14 &47.52 &46.18 &50.11 \\
     \hline
    \textbf{FaVChat} &V &\textbf{70.01} &\textbf{79.79} &\textbf{51.09} &\textbf{62.17} \\ \hline
\end{tabular}
    \end{adjustbox}
\end{wraptable}

In this section, we evaluate \name{} on emotion recognition using two large-scale in-the-wild datasets: DFEW \citep{jiang2020dfew} and MAFW \citep{liu2022mafw}, both designed for dynamic facial expression analysis. DFEW provides a seven-dimensional emotion distribution, while MAFW offers an eleven dimensional distribution.

To make full use of \name’s capabilities, we first prompt FaVChat to generate fine-grained descriptions for each video clip, and then instruct it to infer emotions from these detailed descriptions, enabling more context-aware predictions. As shown in Table~\ref{emotion}, even when using only visual inputs, FaVChat outperforms all existing methods, including models that combine video and audio. This demonstrates the effectiveness of fine-grained descriptions in improving facial emotion recognition. Other zero-shot results are provided in \append\ref{appendix Zero-shot generalizations}.

\subsection{Ablation Study} 
\paragraph{Key Components in \name{}.} To analyze the contributions of key components in \name{}, we conduct ablation experiments on main model variants. 

\begin{wraptable}[6]{r}{6cm}
    \vspace{-2mm}
    \caption{Ablation study results on the impact of key components in \name{}.}
    \label{ablation}
    \begin{adjustbox}{max width=\linewidth}
        \begin{tabular}{lcc}
        \toprule
        Models & DFEW-WAR & MAFW-WAR   \\
        \midrule
        \name{}              &       \textbf{79.79}          &  \textbf{62.17}          \\
        \name{}~ w/o FE              & 55.01               & 55.01        \\
        \name{}~ w/o $CA^2$   & 70.04                 & 57.49           \\
        \name{}~ w/o Adapters &72.53 & 58.31\\
        
        \bottomrule
\end{tabular}
    \end{adjustbox}
\end{wraptable}
\begin{itemize}
\item Without Facial Encoder (w/o FE): We remove the facial encoder and apply the multi-granularity strategy to the general visual encoder, evaluating the necessity of incorporating a dedicated facial encoder.
\item Without $CA^2$ module (w/o $CA^2$): We replace the $CA^2$ module with a standard cross-attention module to assess the contribution of our $CA^2$-based multi-granularity learning approach.
\item Without Weight Adapters: (w/o Adapters): To verify the effectiveness of the Weight  Adapters, we conducted experiments by removing the introduced MoE Adapter component and replacing it with a simple vector weighted summation.
\end{itemize}

We evaluate these variants on DFEW \citep{jiang2020dfew} and MAFW \citep{liu2022mafw}, using the WAR metric as a performance indicator. The results in Table \ref{ablation} show that removing either module leads to a notable performance drop, confirming that both the facial encoder and multi-granularity strategy play essential roles in enhancing fine-grained facial video understanding.

\begin{table*}[!htb]
    \centering
    \small
    \setlength{\tabcolsep}{1.8pt} 
   \begin{adjustbox}{max width=0.78\linewidth} 
    \renewcommand{\arraystretch}{1} 
    \caption{
    Ablation study results on the impact of data-efficient reinforcement learning in \name
    }
    \label{rl_ablation}
    \resizebox{0.98\textwidth}{!}{
    \begin{tabular}{lccccccccc}
        \hline
        \multirow{2}{*}{\textbf{Method}} & \textbf{LLM} & \multicolumn{4}{c}{\textbf{VideoChatGPT Scores}} & \multicolumn{2}{c}{\textbf{N-gram Based}} &\multicolumn{1}{c}{\textbf{Event Based}} \\
        \cmidrule(lr){3-6} \cmidrule(lr){7-8}\cmidrule(lr){9-9}
        & \textbf{Size} &\textbf{Correctness} & \textbf{Detail} & \textbf{Context} & \textbf{Temporal} & \textbf{CIDEr} & \textbf{Rouge-L} & \textbf{AutoDQ} \\
        \hline
        FaVChat$_\mathrm{w.o. RL}$ & 7B & 4.67 & 4.48 & 4.51 & 4.56 & 0.432 & 0.447 & 0.464\\
        FaVChat$_{\mathrm{w.GRPO}}$ & 7B & 4.68 & 4.48 & 4.52 & 4.54 & 0.433 & 0.449 & 0.461 \\
        FaVChat$_{\mathrm{w.GRPO}}$(170K) & 7B & 4.71 & 4.54 & 4.59 & 4.74 & 0.434 & 0.452 & 0.473 \\ 
        FaVChat$_{\mathrm{w.o.RD}}$ & 7B & 4.73 & 4.67 & 4.57 & 4.64 & 0.437 & 0.451 & 0.477 \\
        \textbf{FaVChat$_{\mathrm{DE-GRPO}}$} & 7B &\textbf{4.84} &\textbf{4.79} & \textbf{4.69} &\textbf{4.78} &\textbf{0.443} &\textbf{0.454} &\textbf{0.487} \\
        \hline
        \end{tabular}
    }\end{adjustbox}
    
\end{table*}

\begin{figure}[htbp]
\RawFloats
    \centering
    \begin{minipage}[b]{0.48\textwidth}
    \centering
    \includegraphics[width=\linewidth]{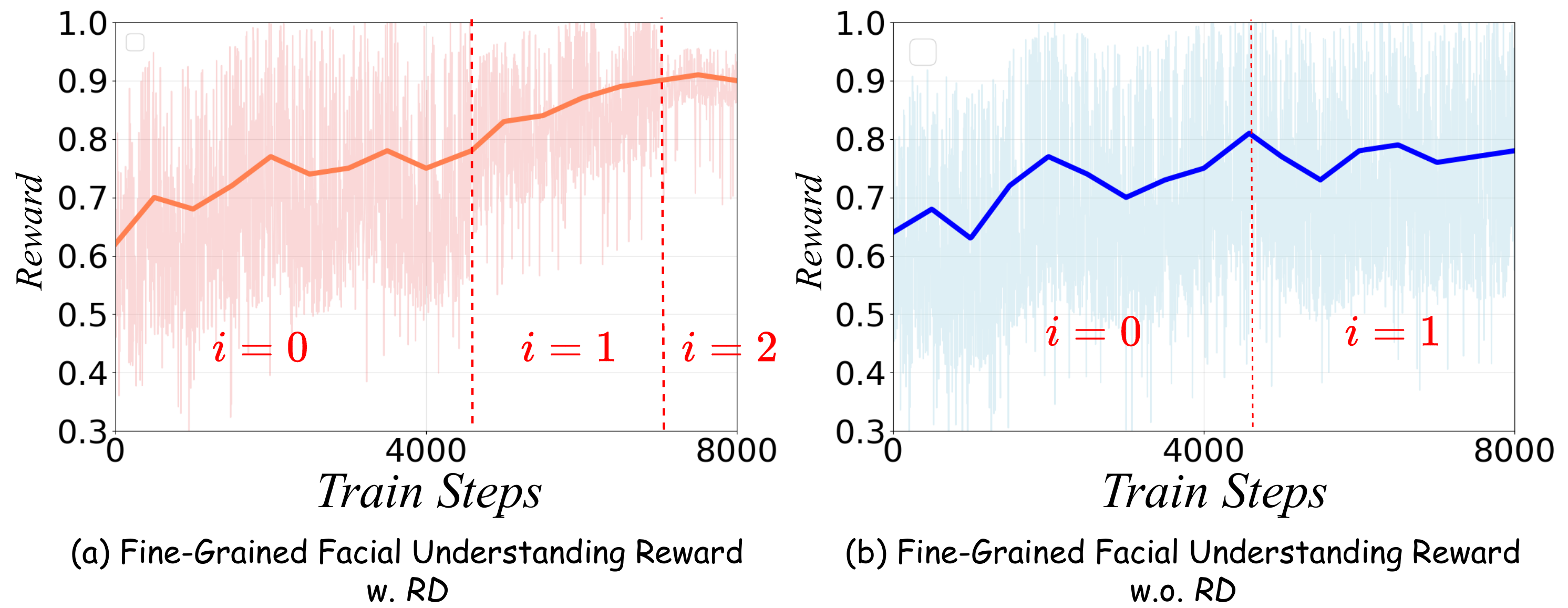}
    \caption{The performance of Facial-Specific Fine-Grained Reward in data recurrent mechanism ablation, where $i$ represents the number of iterations of the training.}
    \label{rl_ablation_visual}
\end{minipage}
\hfill
\begin{minipage}[b]{0.48\textwidth}
    \centering
    \includegraphics[width=\linewidth]{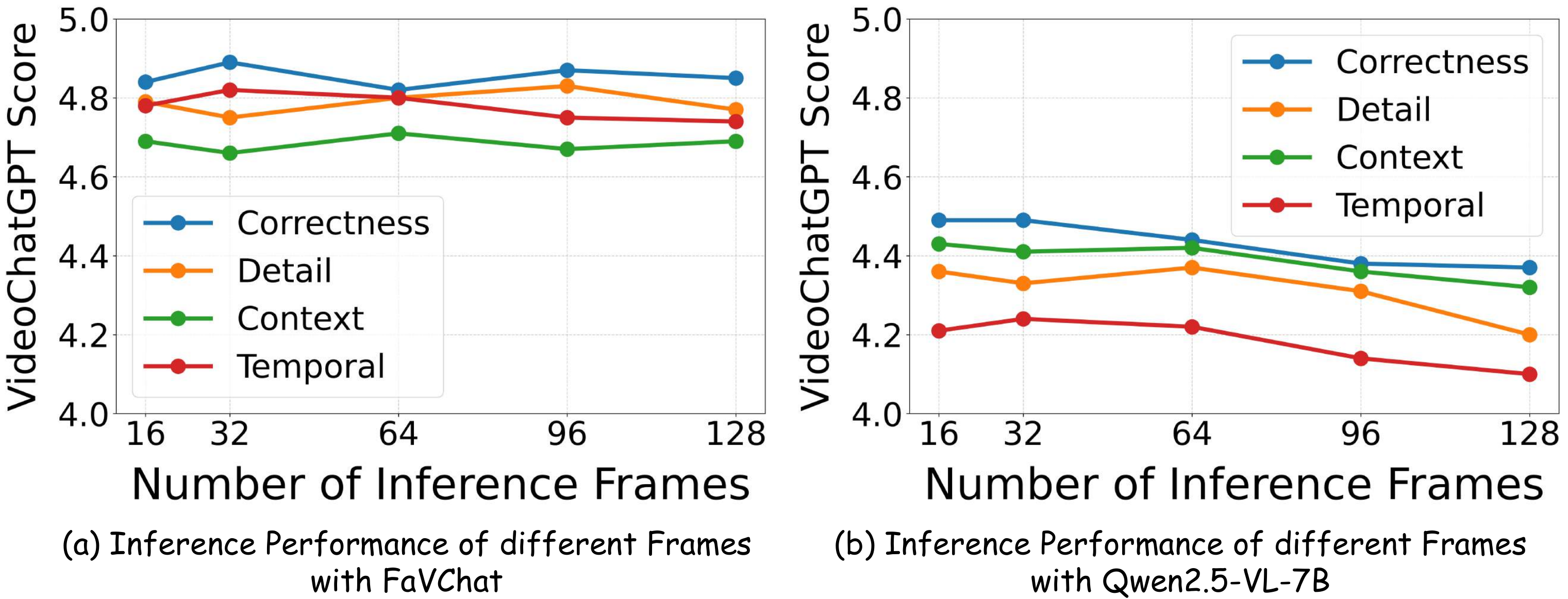}
    \caption{Impact of frame length. While Qwen2.5-VL-7B degrades with long inputs, FaVChat maintains robust performance, demonstrating superior stability.}
    \label{frames_number}
\end{minipage}
\end{figure}

\paragraph{Data-Efficient GRPO.} To verify DE-GRPO during reinforcement learning, we evaluate three variants:

\begin{itemize}[leftmargin =1em, itemsep=0pt, topsep=0pt]
\item Without Reinforcement Learning (w.o. RL): We skip RL and directly evaluate the fine-tuned model to assess whether the RL stage is necessary.
\item With Standard GRPO (w. $\mathrm{GRPO}$): We use standard GRPO rewards, such as output format and length, and train on the full \data{} set to test both our fine-grained reward design and data efficiency.
\item Without Recurrent Data Mechanism (w.o. $\mathrm{{RD}}$): We remove recurrent data scheduling; after the first round, all data are fed into the second round while keeping the total training steps identical to the main setting.
\end{itemize}

All variants are evaluated on DFEC. Table~\ref{rl_ablation} shows that removing any component substantially degrades performance. Fig.~\ref{rl_ablation_visual} shows that recurrent data scheduling steadily improves the fine-grained reward across rounds, whereas removing it yields little improvement after the first round, confirming the contribution of DE-GRPO and its recurrent data mechanism.

\paragraph{Influence of Inference Frame Length.} We evaluate the effect of varying input frame counts during inference on both \name~ and Qwen2.5-VL-7B \cite{Qwen2.5-VL}, using the VideoChatGPT scores on the DEFC benchmark \cite{zhao2025facial} as the evaluation metric. As shown in \fig\ref{frames_number}, \name~ maintains stable performance as the number of input frames increases, whereas Qwen2.5-VL-7B shows a noticeable decline when the input exceeds 96 frames. This demonstrates that \name~ effectively preserves fine-grained information even in long videos, providing a strong basis for our subsequent work on long-video understanding.

\section{Conclusion}


We proposed FaVChat, a multi-modal large language model for fine-grained facial video understanding. With hierarchical prompt-query guided feature extraction and data-efficient GRPO, FaVChat captures subtle facial cues and learns effectively from limited data. Experiments on emotion recognition, explainable reasoning, and facial analysis demonstrate strong performance, detail preservation, and generalization. Future work will extend this approach to other fine-grained video domains and further improve data efficiency.

\bibliographystyle{unsrtnat}
\bibliography{favchat}


\appendix

\newpage

\section{Related Works}
\label{Supplementary:related}

The rapid advancement of multimodal large language models (MLLMs) has spurred the emergence of Video-MLLMs. To mitigate the dramatic increase in visual token count caused by the temporal dimension, existing approaches typically either insert projectors such as MLPs and Q-Formers \cite{li2023blip} after the visual encoder \cite{maaz2023video,xu2024pllava,ma2023vista,li2023videochat,liu2024st} or train a dedicated video encoder from scratch \cite{maaz2023video,luo2023valley,ren2024timechat,wang2024emu3}. However, these strategies often sacrifice fine-grained features particularly in domains with rich and intricate details such as human faces. While significant progress has been achieved in fine-grained facial understanding for static images \cite{narayan2024facexformer,sun2024task,qin2024faceptor,qin2023swinface}, extending such capabilities to videos remains underexplored, with few high-performance frameworks designed specifically for fine-grained video face analysis. This gap represents one of the key challenges that our proposed \name~ addresses. Unlike static face images, facial videos contain additional temporal cues, such as inter-frame facial motion and emotional dynamics, necessitating models that go beyond superficial appearance and capture deeper semantic content. Reinforcement learning (RL) \cite{shao2024deepseekmath} offers a promising avenue to enhance such understanding; however, current label-free preference-based RL methods struggle to supervise fine-grained descriptions \cite{wang2025timezero,zhao2025r1,feng2025video}, while obtaining fine-grained annotations is prohibitively labor-intensive, which is a dual challenge that \name~ seeks to overcome. Below is a more comprehensive discussion of related work.

\subsection{Video Multimodal Large Language Models}

With the rapid advancement of Multimodal Large Language Models (MLLMs), the research community has increasingly shifted its focus towards video. Unlike images, videos introduce an additional temporal dimension, which leads to a multiplicative increase in the number of visual tokens that MLLMs need to process. This complexity has posed significant challenges to the development of Video-MLLMs. On one hand, similar to MLLMs, the projectors of Video-MLLMs are primarily categorized into two types: MLPs and Q-Formers \citep{li2023blip}. MLP-based frameworks directly transform the output of the visual encoder into embeddings. For instance, Video-ChatGPT \citep{maaz2023video} aligns video features with the LLM through linear layers, while PLLaVA \citep{xu2024pllava} employs pooling to mitigate domain differences between images and videos. However, the inherent structure of MLPs limits their capability in handling multi-frame videos. In contrast, the output of Q-Formers is independent of the length of the input video and is instead determined by the number of learnable queries. For example, Vista-LLaMA \citep{ma2023vista} integrates a recursive approach into the application of Q-Former, aiming to address the limited frame differentiation problem of Q-Former. VideoChat \citep{li2023videochat} utilizes additional learnable queries to align video embeddings. ST-LLM \citep{liu2024st} directly applies a pre-trained Q-Former to video frames, resulting in more compact representations. 

On the other hand, since videos can be decomposed into sequences of images, the majority of current research still employs image encoders to extract features from video frames as video representations. For instance, VideoChat \citep{li2023videochat}, Video-ChatGPT \citep{maaz2023video}, Valley \citep{luo2023valley}, and Chat-UniVi \citep{jin2024chat} utilize CLIP ViT \citep{long2015learning} to process videos, while LLaMA-VID \citep{li2024llama}, TimeChat \citep{ren2024timechat}, and Emu3 \citep{wang2024emu3} opt for Eva-clip ViT \citep{sun2023eva} as their visual encoder. However, treating videos merely as sequences of images overlooks the temporal characteristics of videos. Consequently, subsequent researchers advocate employing pre-trained video encoders to capture temporal features. For example, VideoChat2 \citep{li2024mvbench} adopts the pre-trained video foundation model UMT-L \citep{li2023unmasked} to capture video features, and Video-LLaVA \citep{lin2023video}leverages the visual encoder of languageBind \citep{zhu2023languagebind} to process video inputs. Nevertheless, while these methods aim to increase the length of comprehensible videos, they neglect the importance of fine-grained information, resulting in the loss of fine-grained features.

\subsection{Fine-grained Facial Understanding}

The task of fine-grained face understanding aims to capture, parse, and comprehend the detailed features of the human face, and to apply the obtained results to various practical applications. Research in this area based on still images has achieved remarkable performance. Tasks such as face parsing \citep{kuprashevich2023mivolo,te2020edge}, facial attribute recognition \citep{Shu_Yan_Chen_Xue_Shen_Wang_2021,Zheng_Yang_Zhang_Bao_Chen_Huang_Yuan_Chen_Zeng_Wen_2022}, age/gender estimation \citep{cao2020rank,kuprashevich2023mivolo}, head pose estimation \citep{Cobo_Valle_Buenaposada_Baumela_2024,Valle_Buenaposada_Baumela_2021}, and facial expression recognition \citep{chang2018facial,li2020deep} have been extensively studied by the academic community. In addition, subsequent researchers have attempted to build multi-task models to handle multiple tasks using a single general model \citep{Zhang_Zhang_Li_Qiao_2016,Ranjan_Patel_Chellappa_2019,Ranjan_Sankaranarayanan_Castillo_Chellappa_2016}. However, common multi-task models are only capable of performing tasks that are highly correlated. This dilemma has been alleviated with the development of transformers. Face perception models based on transformers, such as FaceXFormer \citep{narayan2024facexformer}, Q-Face \citep{sun2024task}, Faceptor \citep{qin2024faceptor}, and SwinFace \citep{qin2023swinface}, have brought convenience to the research of fine-grained face understanding. Notably, FaceXFormer \citep{narayan2024facexformer} introduced FaceX, an efficient parameterized decoder that provides a universal facial feature representation for multi-tasking. Our proposed FaVChat incorporates FaceXFormer \citep{narayan2024facexformer} as an additional visual encoder, focusing on extracting visual facial features. However, despite the significant progress made in the image domain, there is still a lack of research on high-performance fine-grained video face understanding frameworks when extending the task of fine-grained face understanding to the video domain. This is the focus of the current paper.

\subsection{Reinforcement Learning for Video-MLLMs}

With the remarkable success of multimodal large language models (MLLMs) \cite{deng2025boosting,deng2025openvlthinker,liu2025visual,zhang2025r1}, reinforcement learning has been increasingly extended to video understanding. Early approaches, such as TimeZero \cite{wang2025timezero} and R1-Omni \cite{zhao2025r1}, applied GRPO \cite{shao2024deepseekmath} to video tasks involving temporal distribution modeling and sentiment analysis. Video-R1\cite{feng2025video} enhanced spatial reasoning in videos by integrating spatiotemporal sequential structure into GRPO. Building upon this, STAR-R1 \cite{qistar} introduced a tailored spatiotemporal reward mechanism to further improve the model’s reasoning capabilities in dynamic, long-duration scenarios. In this work, we propose DE-GRPO, a novel reinforcement learning framework specifically designed for data-efficient training with limited high-quality samples. Experimental results show that our method achieves substantial performance gains across a range of face-centric video understanding tasks.

\section{Dataset}
\label{dataset}
\begin{figure*}[!htb]
   \centering
   \includegraphics[width=\linewidth]{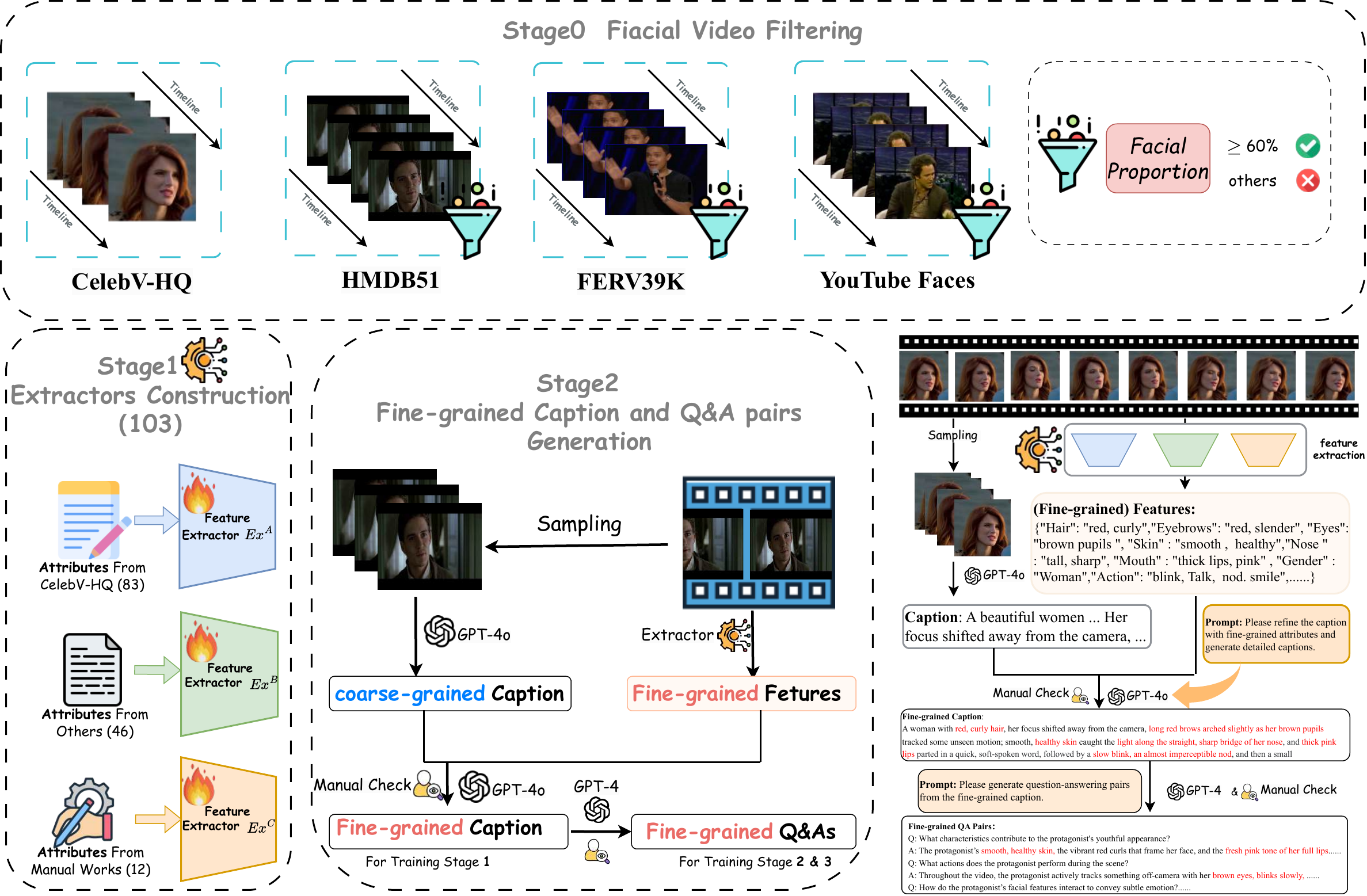}
    \caption{\small
Training Data Creation Process. 
}
    \label{data}
\end{figure*}

\begin{figure*}[h]
   \centering
   \includegraphics[width=0.9\linewidth]{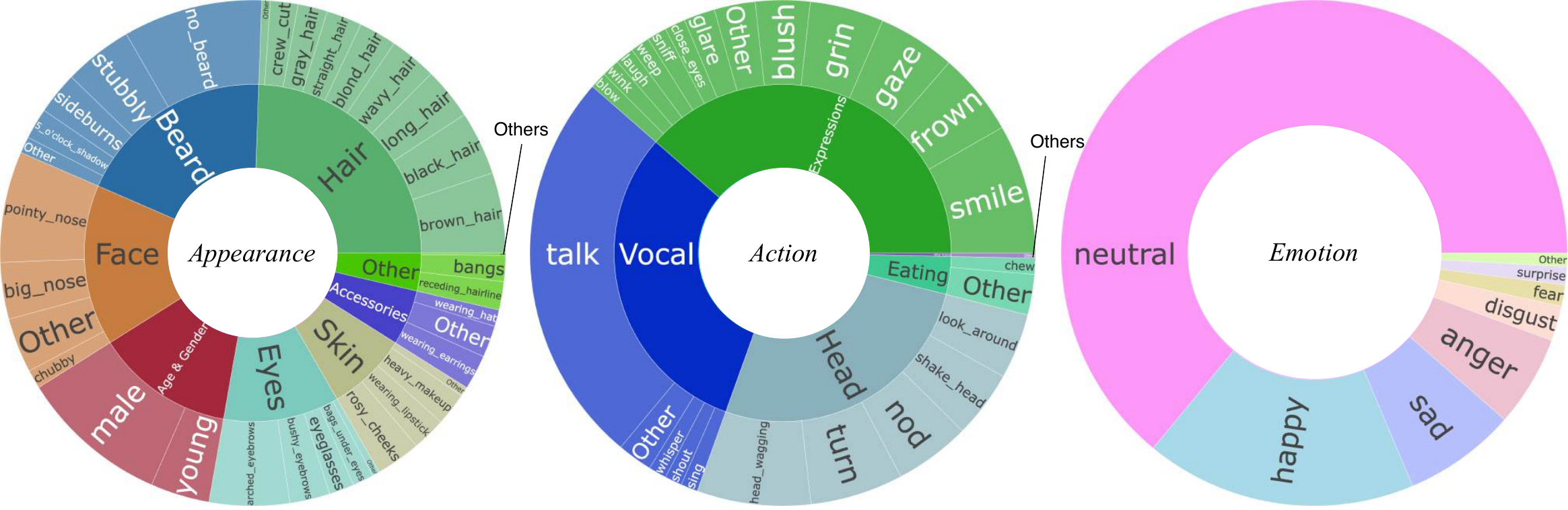}
    \caption{Distribution of the \name~ dataset. For clarity, low-frequency attributes are grouped into an "Other" category.
}
    \label{dataset_distribution}
\end{figure*}

\begin{table*}[h!]
\centering
\begin{adjustbox}{max width=0.98\linewidth}
\begin{tabular}{@{}cccccccccc@{}}
\toprule
\multicolumn{10}{c}{\textbf{Appearance}} \\
\midrule
blurry & male & young & chubby & pale\_skin & rosy\_cheeks & oval\_face & receding\_hairline & bald & bangs \\
610 & 38434 & 15252 & 4880 & 2440 & 11591 & 3660 & 7321 & 1830 & 6711 \\
\midrule
black\_hair & blond\_hair & gray\_hair & brown\_hair & straight\_hair & wavy\_hair & long\_hair & arched\_eyebrows & bushy\_eyebrows & bags\_under\_eyes \\
16472 & 9761 & 7626 & 20742 & 9151 & 12201 & 13422 & 20742 & 10371 & 5491 \\
\midrule
eyeglasses & sunglasses & narrow\_eyes & big\_nose & pointy\_nose & high\_cheekbones & big\_lips & double\_chin & no\_beard & 5\_o'clock\_shadow \\
6467 & 1834 & 1220 & 14621 & 28674 & 3672 & 2928 & 2671 & 35994 & 7981 \\
\midrule
goatee & sideburns & mustache & heavy\_makeup & wearing\_earrings & wearing\_hat & wearing\_lipstick & wearing\_necklace & wearing\_necktie & wearing\_mask \\
976 & 10981 & 2478 & 8541 & 8976 & 4271 & 8663 & 3663 & 3512 & 1021 \\
\midrule
facial\_tattoos & facial\_hair & clean\_shaven & stubbly & shaved\_head & crew\_cut & mullet & bald\_spot &  &  \\
244 & 126 & 421 & 17784 & 276 & 7123 & 62 & 42 &  &  \\
\midrule
\multicolumn{10}{c}{\textbf{Action}} \\
\midrule
blow & chew & close\_eyes & cough & cry & drink & eat & frown & gaze & glare \\
1961 & 1891 & 2441 & 51 & 692 & 1161 & 1432 & 9761 & 9151 & 3678 \\
\midrule
head\_wagging & kiss & laugh & listen\_to\_music & look\_around & make\_a\_face & nod & play\_instrument & read & shake\_head \\
13421 & 918 & 2189 & 1513 & 7688 & 107 & 8907 & 102 & 811 & 7931 \\
\midrule
shout & sign & sing & sleep & smile & smoke & sneeze & sneer & sniff & talk \\
2032 & 1712 & 2001 & 599 & 15241 & 1271 & 22 & 1621 & 2318 & 46775 \\
\midrule
turn & weep & whisper & wink & yawn & blush & grin & grimace & scrunch & squint \\
10981 & 2271 & 2121 & 2098 & 92 & 6421 & 8724 & 102 & 812 & 92 \\
\midrule
stare & smirk & sigh & pout & wince &  &  &  &  &  \\
651 & 61 & 118 & 141 & 271 &  &  &  &  &  \\
\midrule
\multicolumn{10}{c}{\textbf{Emotion}} \\
\midrule
happy & sad & surprise & neutral & anger & contempt & disgust & fear & Shame & Confusion \\
10798 & 4472 & 726 & 39869 & 3629 & 469 & 1403 & 836 & 31 & 36 \\
\toprule

\end{tabular}
\end{adjustbox}
\caption{Attribute statistics results of \name~ dataset. The first row denotes the attribute names, and the second line denotes the occurrence counts of corresponding attributes in the dataset.}
\label{dataset_statistics}
\end{table*}

In this section, we provide a detailed description of the dataset and its construction pipeline. To the best of our knowledge, our dataset is the only face video dataset that includes question-answer annotations and features the largest number of attribute categories. Specifically, it comprises 61,007 high-resolution videos, each annotated with 2 to 4 question-answer pairs and a caption of over 100 tokens. The dataset encompasses 103 attributes, more than any existing face video dataset. \fig\ref{fig:dataset_comparison} compares our dataset with current popular “face-centric” datasets. Additionally, the distribution of attribute counts is shown in Figure \ref{dataset_distribution}, and the attribute statistics are summarized in Table \ref{dataset_statistics}. The construction pipeline consists of three stages: face video filtering, extractor construction, and fine-grained caption and question-answer pair generation. A concrete example is provided in the bottom-right corner of \fig\ref{data}. Detailed descriptions are given in the following subsections.

\subsection{Filtering Facial Videos from Existing datasets}

Our video raw data consists of four parts: CelebV-HQ \citep{zhu2022celebv}, HMDB51 \citep{kuehne2011hmdb}, FERV39K \citep{wang2022ferv39k} and YouTube Faces \citep{DBLP:conf/cvpr/WolfHM11}. Specifically, given the detailed video attributes of CelebV-HQ, we have incorporated all $35,666$ video from CelebV-HQ. For other videos, we first conducted face detaction utilizing AntelopeV2 \citep{ren2023pbidr}. If the minimum proportion of human faces in the video excceded 60\%, the video was considered  to meet the face video standard. Subsequently, based on the average length of videos in CelebV-HQ, we used the OpenCV package to trim the videos, ultimately obtaining an additional $25,341$ video clips, each ranging in length from 36 to 600 frames.

\subsection{Fine-grained Feature Extractors Construction}
\label{Extractors Construction}
To ensure the extraction of a rich set of fine-grained features from each video, we developed a multi-classifier feature extractor. Initially, we sample 1,000 data points from CelebV-HQ and another 1,000 from the combined HMDB51, FERV39K, and YouTube Faces datasets. These samples were used to train two multi-classifiers: $Ex^A$, capable of extracting 83 facial features, and $Ex^B$, which extracts 42 facial features. However, these features still lacked descriptions of subtle facial micro-expressions (e.g., eyebrow raises, frowns, pouts).

To address this, we additionally sample 452 videos from CelebV-HQ, specifically chosen for their pronounced facial actions. These videos underwent meticulous manual annotation for micro-expression features, which then served as training data for a third multi-classifier, $Ex^C$. This classifier was designed to extract 12 distinct micro-expression features. Our comprehensive feature extractor, as illustrated in step 1 of \fig \ref{data}, integrates these three classifiers. After eliminating redundant features, this combined extractor captures a total of 103 facial features. To our knowledge, this represents the largest number of features extracted in the field of facial video analysis, thereby guaranteeing the availability of sufficiently rich, fine-grained labels for our training data.

\subsection{Generating Text for Each Video}
\label{data_draft}

\begin{figure}[!t]
\centering
\begin{minipage}[t]{0.48\textwidth}
\begin{algorithm}[H]
\caption{Weight Adapter Algorithm} 
\label{algorithm_adapter}
\begin{algorithmic}[1] 
\STATE \textbf{Input:} $\mathbf{V} \in \{\mathbf{V}_g,\mathbf{V}_f\} \in \mathbb{R}^{L \times d}$, $\mathbf{T} \in \mathbb{R}^{L \times d}$ \hfill \textcolor{gray}{\textit{Visual and text tokens}}
\STATE \textbf{Initialize:} $W_Q^v \in \mathbb{R}^{d \times d}$, $W_K^t \in \mathbb{R}^{d \times d}$, $W_V^t \in \mathbb{R}^{d \times d}$ \hfill {\textcolor{gray}{\textit{Learnable weights}}}
\STATE $Q_t \gets T W_Q^t$, $K_v \gets V W_K^v$, $V_v \gets V W_V^v$ \hfill {\textcolor{gray}{\textit{Projection to Q, K, V spaces}}}
\STATE $A \gets \text{softmax}\left(\frac{Q_t K_v^\top}{\sqrt{d}}\right)$ \hfill {\textcolor{gray}{\textit{Attention weights}}}
\STATE $H \gets A V_v$ \hfill {\textcolor{gray}{\textit{Weighted aggregation}}}

\STATE $\tilde{V} \gets \text{LayerNorm}(V + H)$ \hfill {\textcolor{gray}{\textit{Residual connection and normalization}}}
\STATE $v_{\text{global}} \gets \frac{1}{L} \sum_{i=1}^{L} \tilde{V}_i$ \hfill {\textcolor{gray}{\textit{Global pooling}}}

\STATE $h_1 \gets \text{GELU}(W_1 v_{\text{global}} + b_1)$, $h_2 \gets \text{GELU}(W_2 h_1 + b_2)$ \hfill {\textcolor{gray}{\textit{MLP layers}}}
\STATE $\text{output} \gets w^\top h_2 + b$ \hfill {\textcolor{gray}{\textit{Final output}}}

\STATE \textbf{Return} $\text{output}$
\end{algorithmic}
\end{algorithm}
\end{minipage}
    \hfill
\begin{minipage}[t]{0.48\textwidth}
\begin{algorithm}[H]
\setstretch{1.11}
\caption{DE-GRPO}
\label{alg:rl}
\begin{algorithmic}[1]
\STATE Initialize $D$, $\theta$, and other hyperparameters
\FOR{$i=0$ \textbf{to} total\_items}
    \STATE Sample batch $B$ from $D$
    \FOR{data $\in B$ }
        \STATE Compute $U^{(i)}$ \hfill \textcolor{gray}{\textit{(Refer to Equation \ref{U_i})}}
        \STATE Update $s^{(i+1)}$ \hfill \textcolor{gray}{\textit{(Refer to Equation \ref{s^i})}}
        \IF{$s^{(i+1)} < \tau_{\text{remove}}$}
            \STATE Remove $i$ from $D_t$
            \STATE \textbf{continue}
        \ENDIF
        \STATE Set $\delta{(i)}$ \hfill \textcolor{gray}{\textit{(Refer to Equation \ref{delta})}}
        \STATE Compute $\bar{A}^{(i)}$ \hfill \textcolor{gray}{\textit{(Refer to Equation \ref{A^i})}}
    \ENDFOR
    \STATE Optimize $L_{\text{SF-RDF}}$ \hfill \textcolor{gray}{\textit{(Refer to Equation \ref{DE-GRPO})}}
\ENDFOR
\end{algorithmic}
\end{algorithm}
\end{minipage}
\end{figure}

We've outlined a detailed process for generating textual descriptions from original videos, as depicted in step 2 of \fig \ref{data}. 
The video textual generation process is further divided into four sub-steps:
\begin{itemize}
    \item {\textbf{Frame Sequence Generation}}. The raw video frames are down sample by 16 to obtain a shorter frame sequence. 
    \item {\textbf{Fine-grained Description Generation}}. We feed the frame sequence from the preceding step to the trained feature extractors. And the output is subsequently converted into a fine-grained feature dictionary via rule-based textual mapping.
    \item {\textbf{Summary Generation}}. we input the frame sequence obtained in the first sub-step and the fine-grained feature dictionary derived in the second sub-step into GPT-4o with an engineered prompt to generate fine-grained summary texts. 
    
    \item {\textbf{QA Pair Generation}}. we employ GPT-4 to generate 4 question-answer (QA) pairs for each summary obtained in the third sub-step. 
\end{itemize}
Note that in sub-steps 3 and 4, we implement manual quality control through sampling during the generation process. Additionally, we conduct multiple rounds of random sampling on the generated text, followed by rigorous human evaluation to ensure that the textual quality meet the stringent requirements for model training.

\section{The complete results of Textual Facial Analysis}
\label{appendix:expression_complete}
In this section, we present all experimental results on the DFEC dataset, summarized in \tab\ref{expression_complete}. The proposed FaVChat model achieves superior performance across all existing VLLMs.

\begin{table*}[!htb]
    \centering
    \small
    \setlength{\tabcolsep}{1.8pt} 
    \caption{\small
    Performance comparison on the DFEC \citep{zhao2025facial} dataset for textual emotion analysis. 
    The $*$ columns show VideoChatGPT Scores (0–10) evaluated on our internally curated test dataset. $\clubsuit$ denotes a model trained to process both visual and speech modalities. "-Face" denotes the variant fine-tuned on our FaVChat-170K dataset. 
    }
    \label{expression_complete}
  \begin{adjustbox}{max width=0.9\linewidth} 
    \begin{tabular}{lcccccccccc}
        \hline
        \multirow{2}{*}{\textbf{Method}} & \textbf{LLM}  & \multicolumn{5}{c}{\textbf{VideoChatGPT Scores}} & \multicolumn{2}{c}{\textbf{N-gram Based}} &\multicolumn{1}{c}{\textbf{Event Based}} \\
        \cmidrule(lr){3-7} \cmidrule(lr){8-9}\cmidrule(lr){10-10}
        & \textbf{Size} &{\textbf{Test(2000)$^*$}} &\textbf{Correctness} & \textbf{Detail} & \textbf{Context} & \textbf{Temporal} & \textbf{CIDEr} & \textbf{Rouge-L} & \textbf{AutoDQ} \\
        \hline
        \rowcolor{gray!20}
        \multicolumn{10}{c}{\textbf{\textit{General VLLM}}} \\ \hline
        VideoLLaMA \citep{zhang2023video}  &7B &6.12 & 3.60 & 3.67 & 3.84 & 3.50 & 0.189 & 0.196 & 0.303  \\
        VideoChat \citep{li2023videochat} &7B &5.92 & 3.47 & 3.52 & 3.92 & 3.38 & 0.251 & 0.192 & 0.344  \\
        VideoChat2 \citep{li2024mvbench}   &7B &6.47 & 3.70 & 3.56 & 4.16 & 3.52 & 0.202 & 0.229 & 0.311  \\
        Chat-UniVI \citep{jin2024chat} &7B &6.76 & 3.64 & 3.63 & 4.21 & 3.61 & 0.189 & 0.231 & 0.396  \\
        LLaVA-Next-Video \citep{zhang2024llava} &7B &7.14 & 4.19 & 4.07 & 4.39 &  4.04 & 0.250 & 0.249 & 0.395  \\
        ShareGPT4Video \citep{chen2024sharegpt4video} &7B & 7.33& 4.24 & 4.13 & 4.35 &  4.09 & 0.192 & 0.205 & 0.394  \\
        LLaMA-VID \citep{li2024llama} &7B & 6.98& 3.95 & 4.01 & 4.22 & 3.71 & 0.195 & 0.231 & 0.339  \\
        VideoLLaMA2 \citep{cheng2024videollama} &7B &7.41 & 4.17 & 4.02 & 4.47 & 3.93 & 0.253 & 0.266 & 0.344  \\
        PLLaVA \citep{xu2024pllava} &7B & 7.47 & 4.21 & 4.15 & 4.37 & 4.08 & 0.268 & 0.250 & 0.393  \\
        ST-LLM \citep{liu2024st} &7B & 6.79& 4.00 & 3.98 & 4.31 & 3.94 & 0.213 & 0.238 & 0.321  \\
        Tarsier \citep{wang2024tarsier} &7B  &6.42 & 3.59 & 3.50 & 4.07 & 3.41 & 0.143 & 0.185 & 0.415  \\
        LLaVA-OneVision \citep{li2024llava} &7B &6.72 & 3.68 & 3.47 & 4.10 & 3.42 & 0.115 & 0.165 & 0.379  \\
        Qwen2.5-VL-7B \citep{Qwen2.5-VL} &7B  & 7.84 & 4.49 & 4.36 & 4.43 & 4.21 & 0.261 & 0.254 & 0.417  \\
        Qwen2.5-VL-72B \citep{Qwen2.5-VL} &72B & 8.47 &4.61 & 4.47 & 4.59 & 4.78 & 0.281 & 0.317 & 0.453 \\
        Qwen2.5-VL-3B \citep{Qwen2.5-VL} &2B & 7.01 &4.01 &3.98 &4.37 &3.88 &0.202 &0.221 &0.406 \\
        Qwen3-VL-8B \cite{bai2025qwen3vltechnicalreport} & 8B & 7.99 & 4.49 & 4.37 & 4.41 & 4.33 & 0.262 & 0.257 & 0.421 \\
        Qwen3-VL-30B-A3B \cite{bai2025qwen3vltechnicalreport} & 30B & 8.44 & 4.63 & 4.51 & 4.57 & 4.81 & 0.304 & 0.331 & 0.461 \\
        Claude4-Sonnet \citep{Claude2025} & - & 7.84 &4.55 &4.64 &4.77 &4.84 &0.312 &0.342 &0.472\\
        GPT-4o \citep{openai2024gpt4o} & - & 7.67 &4.22 &3.97 &4.48 & 3.90 & 0.264 &0.213 &0.432 \\
        \hline
         \rowcolor{gray!20}
        \multicolumn{10}{c}{\textbf{\textit{Human-Centric VLLM}}} \\ 
        \hline
        Face-MLLM \cite{sun2024face} & 7B & 6.62 & 3.31 & 3.54 & 3.79 & 2.67 & 0.204 & 0.197 & 0.304 \\
        FaceTrack-MM \cite{zhao2025facial} & 7B & 7.47 & 4.42 & 4.30 & 4.60 & 4.26 & 0.418 & 0.473 & 0.483 \\
        Qwen2.5-VL-7B-Face \cite{Qwen2.5-VL} & 7B & 8.64 & 4.51 & 4.41 & 4.45 & 4.22 & 0.274 & 0.261 & 0.463 \\
        Qwen3-VL-8B-Face \cite{bai2025qwen3vltechnicalreport} & 8B & 8.87 & 4.58 & 4.46 & 4.47 & 4.51 & 0.297 & 0.312 & 0.429 \\
        Emotion-LLaMA$^\clubsuit$ \cite{cheng2025emotion} & 7B & 7.42 & 4.17 & 4.33 & 4.54 & 4.19 & 0.294 & 0.413 & 0.372\\
        HumanOmni$^\clubsuit$ \cite{zhao2025humanomni} & 7B & 7.51 & 4.58 & 4.41 & 4.70 & 4.41 & 0.412 & 0.468 & \textbf{0.523} \\
        Omni-Emotion$^\clubsuit$ \cite{yang2025omniin} & 7B & 7.53 & 4.57 & 4.38 & 4.73 & 4.24 & 0.407 & 0.461 & 0.504 \\
        \cmidrule(lr){1-10} 
        \textbf{FaVChat$_{\mathrm{Qwen2.5}}$} & 7B & 8.89 &4.84 &4.79 & 4.69 &4.78 &0.443 &0.454 &0.487 \\
        \textbf{FaVChat$_{\mathrm{Qwen3}}$} & 8B & \textbf{9.01} & \textbf{4.87} & \textbf{4.82} & \textbf{4.74} & \textbf{4.81} &\textbf{0.447} & \textbf{0.469} & \textbf{0.523} \\
        \hline
    \end{tabular}
    \end{adjustbox}
    
\end{table*}

\section{Statement on Ethics, Privacy, and Bias}
\label{appendix:ethics}

All original facial videos in our proposed FaVChat-170K dataset are sourced from open-source video datasets that have undergone rigorous review and formal ethical risk assessment. Additionally, during the generation of summaries and question-answer pairs, we rigorously monitored and regulated the model’s generation process to mitigate ethical risks—specifically discrimination related to sociodemographic factors, race, or gender—thereby ensuring compliance with established academic ethics standards.

\section{Adapter Supplement}
\label{adapter}

This section serves as a supplement to Section \sec \ref{high_level} of the main text where we formally describe \name’s high-level prompt-query learning. Specifically, two independent learnable weight adapters integrate visual features and prompt-query textual features, producing corresponding learnable scalars. We provide the associated pseudocode implementation in Algorithm \ref{algorithm_adapter}.

The two independent weight adapters employ the same training strategy and produce separate scalar outputs, which are then normalized using softmax to generate complementary weighting coefficients.

\section{RL Supplement}
\label{appendix:rl}

\begin{wraptable}{r}{8.0cm}
    \vspace{-1mm}
    \caption{Statistical results of hyperparameters appearing in DE-GRPO}
    \label{params:rl}
    \begin{adjustbox}{max width=0.95\linewidth}
        \begin{tabular}{@{}clc@{}}
\toprule
\textbf{Symbol} & \textbf{Meaning} & \textbf{Value in \name} \\ \midrule
$\lambda$ & Recurrent smoothing & 0.8 \\
$\tau_{\text{remove}}$ & Remove threshold & 0.20 \\
$\tau_{\text{keep}}$ & Keep threshold & 0.80 \\
$\delta$ & Reward decay factor & 0.50 \\
$\epsilon$ & PPO clip & 0.20 \\
$\beta$ & KL coefficient & 0.01 \\ \bottomrule
\end{tabular}
    \end{adjustbox}
\end{wraptable}
In \sec \ref{method:rl} of the main text, the data recurrent mechanism in DE-GRPO, which is grounded in data efficiency, is introduced, and the reinforcement learning advantage function $\overline{\boldsymbol{A}}^{(i)}$ derived from this mechanism is described. We now provide a complete specification of the DE-GRPO procedure.

After obtaining the advantage function based on the data recurrent mechanism, we substitute it into the GRPO formulation. First, we define the following:

\begin{equation}
\small
\rho_w^{(i)}=\frac{\pi_\theta(y_w^{(i)}|x^{(i)})}{\pi_{\mathrm{old}}(y_w^{(i)}|x^{(i)})},\quad\rho_l^{(i)}=\frac{\pi_\theta(y_l^{(i)}|x^{(i)})}{\pi_{\mathrm{old}}(y_l^{(i)}|x^{(i)})},
\end{equation}

\begin{equation}
    \small
    \quad\Delta^{(i)}=\rho_w^{(i)}-\rho_l^{(i)},
\end{equation}

where $i$ denotes the iteration index of the data recurrent mechanism, and the overall objective function of DE-GRPO is as follows:

\begin{equation}
\begin{aligned}
\label{DE-GRPO}
\small
\mathcal{J}_{\mathrm{DE}-GRPO} = \\
\mathbb{E}_{\mathcal{B}_t}\Bigg[ & \min\left(\log\sigma(\Delta^{(i)}) \cdot \bar{A}^{(i)}, \right. \left. \log\sigma(\mathrm{clip}(\Delta^{(i)}, 1-\epsilon, 1+\epsilon)) \cdot \bar{A}^{(i)}\right) + \beta\operatorname{KL}[\pi_\theta\|\pi_{\mathrm{ref}}] \Bigg]
\end{aligned}
\end{equation}

We present the complete DE-GRPO procedure in Algorithm \ref{alg:rl} and summarize the associated parameters in Table \ref{params:rl}.

\paragraph{Rationale for the Decay Factor Design.}
In Equation \ref{delta}, the decay factor $\delta^{(i)}$ is designed to govern the dynamic lifecycle of training samples and prevent overfitting.
Intuitively, one might expect high-utility samples (where $s^{(i)} > \tau_{keep}$) to be retained with a high weight.
However, if we assign a factor $\ge 1$ to these samples, their accumulated score $s^{(i)}$ would monotonically increase, causing them to permanently dominate the training batch.
This would lead to two issues: (1) the model would overfit to these specific samples, and (2) new, potentially informative samples would be blocked from entering the active training set.
Therefore, we set $\delta < 1$ (e.g., 0.5) specifically when $s^{(i)}$ exceeds the upper threshold $\tau_{keep}$.
This mechanism acts as a ``cooling-down'' strategy: once a sample has been identified as high-utility and utilized for training, we explicitly reduce its priority to reflect the diminishing marginal returns of repeated learning.
Conversely, for samples in the intermediate range ($\tau_{remove} \le s^{(i)} \le \tau_{keep}$), we set $\delta(i) = 1$ to allow them to accumulate utility scores based on their merit until they either prove useful enough to trigger the training threshold or fade out to be discarded.
This design ensures a continuous flow of data where samples are discovered, exploited, and eventually deprecated. 

\paragraph{Clarification on Sample-Level Advantage.}
A critical distinction between standard GRPO \cite{shao2024deepseekmath} and our proposed DE-GRPO lies in the granularity of the advantage function.
In standard GRPO, the advantage is computed per response to estimate the relative quality of each generated output against the group baseline.
In contrast, DE-GRPO aims to optimize data efficiency by prioritizing high-utility training samples.
Consequently, the advantage $\overline{A}^{(i)}$ in Equation \ref{A^i} is computed as an average scalar shared across all response pairs $\mathcal{P}$ derived from the same input video $x^{(i)}$.
Here, $\overline{A}^{(i)}$ functions as a \textit{sample-level importance weight} rather than a \textit{response-level alignment signal}.
It scales the gradient updates based on the sample's current utility: inputs that yield highly separable rewards (indicating rich learning signals) are assigned larger advantage values, thereby amplifying their contribution to the policy update.
This design ensures that the model focuses its capacity on learning from the most informative video samples in each iteration.

\paragraph{Why Geometric Mean for Reward Separability and Gradient Sensitivity.}
In Equations~\ref{U_i}'s two utility components, we aggregate pair-wise quantities using the geometric mean rather than the arithmetic mean:
\begin{equation}
\small
\hat R^{(i)}=\left(\prod_{p=1}^{|\mathcal{P}|} d^{(i)}_{R,p}\right)^{\frac{1}{|\mathcal{P}|}},
\quad
\hat g^{(i)}=\left(\prod_{p=1}^{|\mathcal{P}|} d^{(i)}_{g,p}\right)^{\frac{1}{|\mathcal{P}|}},
\end{equation}
where $d^{(i)}_{R,p}=|\boldsymbol{R}(y_w^{(i)})-\boldsymbol{R}(y_l^{(i)})|$ and $d^{(i)}_{g,p}=\|g(y_w^{(i)})-g(y_l^{(i)})\|_2$ for the $p$-th preference pair.
The rationale follows the robustness observation in Geometric-Mean Policy Optimization~\citep{zhao2025geometricmean}: arithmetic averaging can be dominated by a few outlier terms, while geometric averaging suppresses their influence and yields more stable optimization signals.
This property is particularly important in DE-GRPO because candidate responses for the same facial video can contain occasional abnormal reward gaps or gradient deviations caused by hallucinated facial attributes, ambiguous expressions, or noisy pair construction.
If we used the arithmetic mean,
\begin{equation}
\small
\bar d_{\mathrm{arith}}=\frac{1}{|\mathcal{P}|}\sum_{p=1}^{|\mathcal{P}|} d_p,
\end{equation}
a single extreme $d_p$ could make the whole sample appear highly useful, even if most response pairs provide weak or inconsistent supervision.
By contrast, the geometric mean
\begin{equation}
\small
\bar d_{\mathrm{geo}}=\left(\prod_{p=1}^{|\mathcal{P}|}d_p\right)^{\frac{1}{|\mathcal{P}|}}
=\exp\left(\frac{1}{|\mathcal{P}|}\sum_{p=1}^{|\mathcal{P}|}\log d_p\right)
\end{equation}
aggregates evidence in log space, so an unusually large value contributes only logarithmically.
For positive $d_p$, the AM-GM inequality gives
\begin{equation}
\small
\left(\prod_{p=1}^{|\mathcal{P}|}d_p\right)^{\frac{1}{|\mathcal{P}|}}
\leq
\frac{1}{|\mathcal{P}|}\sum_{p=1}^{|\mathcal{P}|} d_p,
\end{equation}
which means that the geometric mean has a narrower value range than the arithmetic mean.
Therefore, a sample obtains a high utility score only when most preference pairs consistently exhibit clear reward separability and gradient sensitivity.
This design prevents DE-GRPO from repeatedly selecting samples due to isolated outlier pairs and makes the recurrent data mechanism more stable.

\paragraph{Why Using an $L_2$ Norm in Gradient Sensitivity but not in Reward Separability.}
Equation~\ref{U_i} measures two different objects: reward separability compares scalar rewards, whereas gradient sensitivity compares vector-valued update directions.
For the reward term, $\boldsymbol{R}(y_w^{(i)})-\boldsymbol{R}(y_l^{(i)})$ is already a scalar.
Taking its absolute value is sufficient:
\begin{equation}
\small
d_R^{(i)}=\left|\boldsymbol{R}(y_w^{(i)})-\boldsymbol{R}(y_l^{(i)})\right|.
\end{equation}
Applying an $L_2$ norm here would be mathematically redundant, since for any scalar $a$,
\begin{equation}
\small
\|a\|_2=\sqrt{a^2}=|a|.
\end{equation}
Thus, Equation \ref{eq_8} does not omit an additional normalization; its absolute value is already the scalar form of an $L_2$ distance.

In contrast, the gradient term compares two parameter-space update signals:
\begin{equation}
\small
g(y)=\nabla_{\theta}\log \pi_{\theta}(y\mid x), \quad
\Delta g^{(i)}=g(y_w^{(i)})-g(y_l^{(i)}).
\end{equation}
The analysis of positive and negative reinforcement dynamics~\citep{zhu2025negative} shows that reward signals shape learning through their induced gradients on logits and policy parameters; hence, for data utility, the relevant question is not only whether two responses have different rewards, but whether they would push the policy in meaningfully different directions.
To convert the vector $\Delta g^{(i)}$ into a scalar sensitivity score, we use the Euclidean norm:
\begin{equation}
\small
d_g^{(i)}=\|\Delta g^{(i)}\|_2
=\sqrt{\sum_{m=1}^{|\theta|}\left(g_m(y_w^{(i)})-g_m(y_l^{(i)})\right)^2}.
\end{equation}
This choice has three advantages.
First, it is sign-invariant, so opposite coordinate directions do not cancel each other as they would under a raw vector sum.
Second, it measures the actual parameter-space distance between two first-order update directions: with learning rate $\eta$, the distance between the two hypothetical updates is
\begin{equation}
\small
\|\eta g(y_w^{(i)})-\eta g(y_l^{(i)})\|_2
=\eta\|g(y_w^{(i)})-g(y_l^{(i)})\|_2.
\end{equation}
Third, compared with using only reward gaps, the $L_2$ norm captures whether winning and losing responses induce substantially different optimization behavior.
This is aligned with gradient-based analyses of RL objectives, where clipping, KL regularization, and advantages may rescale or dampen gradients but the core learning signal remains expressed through gradient direction and magnitude~\citep{zhu2025negative}.
Therefore, Equation~9 uses an $L_2$ norm because gradient sensitivity is intrinsically vector-valued, while Equation~8 uses an absolute scalar reward difference because reward separability is already one-dimensional.

\section{More Ablations}

\subsection{Impact of Different Train Stages in Hierarchical Prompt-Query Learning}
\label{appendix: Hierarchical Prompt-Query Learning}

\begin{wraptable}{r}{8.0cm}
    \vspace{-1mm}
    \caption{Ablation study results on hierarchical prompt-query learning in FaVChat}
    \label{hierarchical prompt-query learning}
    \begin{adjustbox}{max width=0.95\linewidth}
        \begin{tabular}{@{}lccc@{}}
\toprule
\textbf{model} & \textbf{Test$_{low}$} & \textbf{Test$_{mid}$}  & \textbf{Test(2,000)$_{high}$} \\ \midrule
\name{}$_\mathrm{Stage1}$ & 8.13 & 6.14 & 5.18 \\
\name{}$_\mathrm{Stage2}$ & 8.67 & 8.23(2.09 $\uparrow$) & 6.37  \\
\name{}$_\mathrm{Stage3}$ & 8.91 & 8.44 & 8.31(1.94 $\uparrow$) \\
\name{} & - & - & 8.89 \\
\bottomrule
\end{tabular}
    \end{adjustbox}
\end{wraptable}
As discussed in \sec\ref{train stratege}, the distinct training stages of \name{} correspond to feature learning at progressively higher levels of abstraction. To validate the effectiveness of our hierarchical prompt-guided feature learning approach, we partitioned the internal dataset (introduced in \sec\ref{Quantitative Comparison}) into three test sets reflecting increasing feature levels. Specifically, the low-level test set comprises cases involving only basic facial textures (e.g., skin, hair). The mid-level test set extends this by incorporating cases covering fine-grained facial regions (e.g., eyes, nose, mouth). The high-level test set retains the complete in-house test set of 2,000 samples. We then evaluate \name{} at each training stage across all three test sets to assess the developmental trajectory of feature representation learning. The quantitative results are summarized in Table \ref{hierarchical prompt-query learning}. Experimental results demonstrate that the performance on the internal dataset progressively improves as training proceeds through successive stages. Specifically, FaVChat in Stage 1 excels only on the low-level test set, while Stage 2 exhibits a significant improvement on the medium-level test set, and Stage 3 achieves notable performance even on the high-level test set. This progression validates the effectiveness of our hierarchical feature learning approach.

\subsection{Parameter Analysis}
\label{Parameter Analysis}

\paragraph{Adaptive Adapter Allocation.}
\begin{wrapfigure}[17]{r}{0.53\linewidth}
    \centering
    \includegraphics[width=\linewidth]{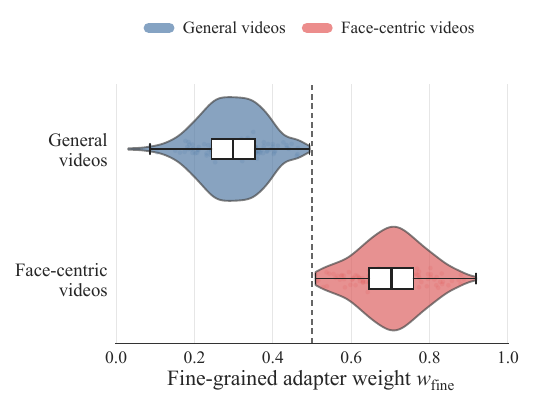}
    \caption{Distribution of fine-grained adapter weights on general and face-centric videos.}
    \label{fig:adapter_weight_allocation}
    \vspace{-4mm}
\end{wrapfigure}
We further analyze whether \name{} dynamically assigns different adapter weights according to the input domain. Specifically, we randomly sample 500 general videos from MVBench \citep{li2024mvbench} and VideoMME \citep{fu2025video}, and 500 face-centric videos from the FaVChat test set. For each video, we record the normalized weights assigned to the general adapter and the fine-grained facial adapter, where the two weights sum to one. Therefore, we visualize the fine-grained adapter weight $w_{\mathrm{fine}}$ as an indicator of adapter allocation preference. As shown in Fig.~\ref{fig:adapter_weight_allocation}, general videos exhibit lower $w_{\mathrm{fine}}$ values and rely more on the general adapter, while face-centric videos are clearly shifted toward higher $w_{\mathrm{fine}}$ values, indicating stronger activation of the fine-grained facial adapter. This result suggests that \name{} does not use a fixed adapter mixture, but adaptively routes different video domains to the corresponding adapter capacity.

\begin{figure}[!htb]
    \centering
    \includegraphics[width=0.88\linewidth]{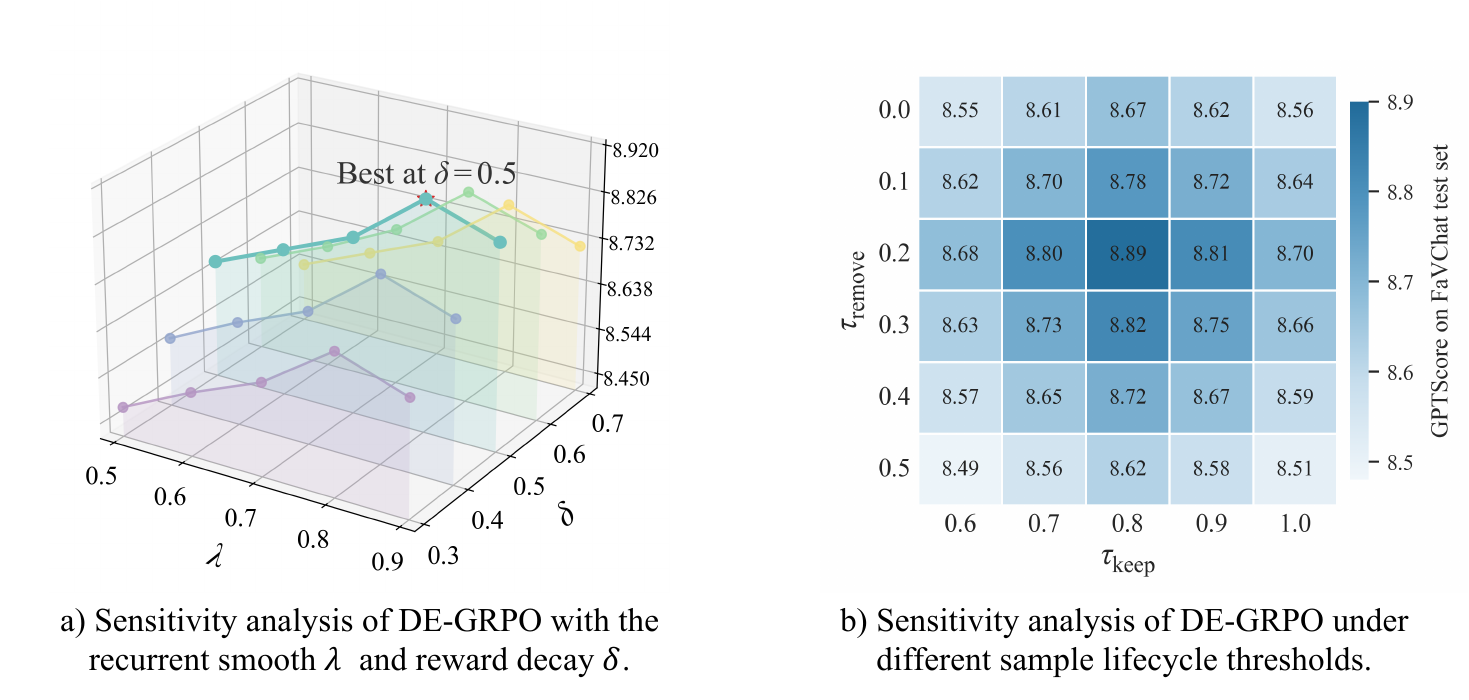}
    \caption{Hyperparameter analysis of DE-GRPO}
    \label{fig:de_grpo_hyperparameter_analysis}
\end{figure}

\paragraph{Hyperparameters in DE-GRPO.}
For the hyperparameters summarized in Table~\ref{params:rl}, the PPO clip coefficient $\epsilon$ and KL coefficient $\beta$ follow commonly used settings in prior GRPO-style reinforcement learning literature~\citep{shao2024deepseekmath,guo2025deepseek}. Therefore, our parameter analysis mainly focuses on the DE-GRPO-specific recurrent data mechanism, including the recurrent smoothing coefficient $\lambda$, the reward decay factor $\delta$, and the sample lifecycle thresholds $\tau_{\mathrm{remove}}$ and $\tau_{\mathrm{keep}}$. As shown in Fig.~\ref{fig:de_grpo_hyperparameter_analysis}(a), the best performance is achieved when $\lambda=0.8$ and $\delta=0.5$, reaching a GPTScore of $8.89$ on the FaVChat test set. A smaller $\lambda$ makes the sample state overly sensitive to short-term utility fluctuations, while an overly large $\lambda$ slows down lifecycle updates. Similarly, an excessively small $\delta$ suppresses informative samples too aggressively after they are selected, whereas a large $\delta$ weakens the cooling-down effect and may repeatedly optimize already exploited samples. Fig.~\ref{fig:de_grpo_hyperparameter_analysis}(b) further analyzes the lifecycle thresholds, where $\tau_{\mathrm{remove}}=0.2$ and $\tau_{\mathrm{keep}}=0.8$ obtain the best result. A low removal threshold retains low-utility or noisy samples for too long, while a high removal threshold prematurely discards useful samples; meanwhile, a low keep threshold cools samples before sufficient exploitation, whereas a high keep threshold delays decay and reduces data turnover. These trends support the default DE-GRPO configuration used in \name{}, which balances stable utility estimation, sufficient exploitation, and continuous sample replacement.

\section{Implementation Details}
\label{Implementation Details}

We use Qwen2.5-7B \citep{qwen2,qwen2.5} as the backbone LLM, with UMT-L \citep{li2023videochat,li2024mvbench} as the general visual encoder and the FaceXFormer feature extractor \citep{narayan2024facexformer} as the facial encoder. A lightweight Q-Former \citep{li2023blip} together with two linear layers serves as the projection module. All experiments are conducted on 8 NVIDIA A100 GPUs. In the three-stage fine-tuning pipeline, we use a learning rate of $2\times10^{-4}$ with a batch size of 16 in the first stage, keep the same learning rate but reduce the batch size to 8 in the second stage, and adopt a learning rate of $2\times10^{-5}$ with a batch size of 8 in the final stage. During reinforcement learning, we further lower the learning rate to $2\times10^{-6}$, use a batch size of 8, and set the weight decay to $0.8$.

\section{More Zero-shot Generalizations}
\label{appendix Zero-shot generalizations}

\begin{wraptable}{r}{8.0cm}
    \caption{Comparisons of different methods on the facial recognition task. ``w. \name{}'' denotes that fine-grained facial features extracted by \name{} are provided as additional inputs to MLLMs, enhancing their decision-making process.}
    \label{recognition}
    \begin{adjustbox}{max width=0.95\linewidth}
        \begin{tabular}{l|ccc}
    \hline
               Methods  &Size & Acc & Acc$_{ w. \mathrm{FaVChat}}$  \\ \hline
               GPT-4o \citep{openai2024gpt4o} & - & 71.29 &93.61\\ 
               Claude4-Sonnet \citep{Claude2025} & - & 73.41 &\textbf{96.14}\\
               VideoChat \citep{li2023videochat} & 7B &54.74 & 73.61\\
               VideoChat2 \citep{li2024mvbench} & 7B & 57.30& 81.32\\
               VideoLLaMa2 \citep{cheng2024videollama} & 7B & 51.03& 75.05\\
               Qwen2.5-VL-7B-Face \citep{Qwen2.5-VL} & 7B &64.28 & 85.58\\
               Qwen2.5-VL-72B \citep{Qwen2.5-VL} &72B &66.84 & 85.87\\ \hline
               \textbf{\name{}} & 7B &\textbf{88.38} & - \\
               \hline
    \end{tabular}
    \end{adjustbox}
\end{wraptable}

\paragraph{Face Recognition.} Face recognition can be viewed as a matching task in which two facial video clips are compared to determine whether they belong to the same individual. The YouTube Faces dataset is well suited for this setting, as it contains multiple videos per person under diverse conditions. Using this dataset, we construct a test set of 3,190 video segments from YouTube Faces \citep{DBLP:conf/cvpr/WolfHM11} that are excluded from training. Each segment contains 36 frames and covers a total of 1,595 unique identities. We evaluate this test set on several models, including GPT-4o \citep{hurst2024gpt}, Claude-4-Sonnet \citep{Claude2025}, VideoChat \citep{li2023videochat}, VideoChat2 \citep{li2024mvbench}, Qwen2.5-VL (2B, 7B, 72B) \citep{wang2024qwen2}, and VideoLLaMA2 \citep{cheng2024videollama}, using accuracy as the metric. As shown in Table \ref{recognition}, \name~ achieves state of the art performance on the facial recognition task. Moreover, when fine grained facial features extracted by \name~ are used as inputs to other models, all of them exhibit clear performance gains. This further highlights \name’s strong ability to capture fine grained facial information, which is crucial for accurate face recognition.

\newpage
\paragraph{Explainable Multimodal Emotion Reasoning.}

\begin{wraptable}{r}{8.0cm}
    \vspace{-1mm}
    \caption{Comparison of explainable multimodal emotion reasoning performance on the EMER dataset \citep{lian2023explainable}.}
    \label{EMER}
    \begin{adjustbox}{max width=0.98\linewidth} 
\begin{tabular}{lcc}
\\
            \toprule
            Models & Clue Overlap & Label Overlap\\
            \midrule
            Ground Truth  & {10.0} &{10.0} \\ 
            \midrule
            VideoChat-Text \citep{li2023videochat}      & {6.32}    & {3.97}\\
            Video-LLaMA \citep{zhang2023video}         & {6.51}    & {4.72}\\
            Video-ChatGPT \citep{maaz2023video}       & {6.98}    & {5.79}\\
            PandaGPT \citep{su2023pandagpt}            & {7.15}    & {5.49}\\
            VideoChat-Embed \citep{li2023videochat}     & {7.15}    & {5.65}\\
            Valley \citep{luo2023valley}              & {7.31}    & {5.76}\\
            \midrule
            \textbf{\name{}} & \textbf{8.41} & \textbf{7.01}\\
            \bottomrule
\end{tabular}\end{adjustbox}
\end{wraptable}
The creation of the EMER dataset \citep{lian2023explainable} highlights an important insight for the community: when additional multimodal emotional cues, such as subtle facial movements, contextual video information, and emotional categories, are incorporated, interpretable emotional reasoning becomes more critical than mere emotion recognition. To evaluate this aspect, we assess \name{} on the Explainable Multimodal Emotional Reasoning (EMER) task. Following prior works, we employ GPT-4o for evaluation, measuring performance using two key metrics: (1) overlap of relevant cues and (2) overlap of final emotional states. The results, shown in Table \ref{EMER}, demonstrate that \name{} consistently outperforms existing models on both metrics, validating its ability to leverage fine-grained multimodal cues for enhanced emotional reasoning.
\paragraph{General and Fine-Grained Video Reasoning.}
To further examine whether facial-domain specialization affects the general video understanding capability of the backbone model, we evaluate \name{} on four zero-shot video reasoning benchmarks, including MVBench \citep{li2024mvbench}, VideoMME \citep{fu2025video}(for general video reasoning), MotionBench \citep{hong2025motionbench}, and UFVideoBench \citep{pan2025ufvideo}(for fine-grained video reasoning). These benchmarks cover both general video reasoning and fine-grained temporal reasoning, providing a complementary assessment beyond face-centric tasks. As shown in Table \ref{tab:general_and_fine_grained_video_reasoning}, \name{} maintains comparable or slightly better performance than its base model Qwen2.5-VL-7B across all evaluated settings. Specifically, \name{} improves the scores from 69.6 to 70.4 on MVBench and from 53.1 to 54.2 on VideoMME. On fine-grained video reasoning benchmarks, it also improves MotionBench from 53.0 / 48.8 to 53.9 / 50.1 and UFVideoBench from 2.54 / 3.99 / 3.78 to 2.59 / 4.07 / 3.84. These results indicate that the proposed facial-domain training does not lead to noticeable forgetting of general video knowledge. Instead, with the adapter-based design and staged optimization, \name{} preserves the broad video reasoning ability of the underlying multimodal model while acquiring stronger fine-grained facial video understanding.

\begin{table*}[!htb]
    \centering
    \small
    \setlength{\tabcolsep}{2.5pt}
    \renewcommand{\arraystretch}{1.1}
    \caption{Zero-shot generalization results on general and fine-grained video reasoning benchmarks. \name{} preserves the general video reasoning capability of Qwen2.5-VL-7B after facial-domain specialization and achieves slight improvements across all evaluated benchmarks.}
    \label{tab:general_and_fine_grained_video_reasoning}
    \begin{adjustbox}{max width=\linewidth}
        \begin{tabular}{@{}lcccc@{}}
\toprule
\rowcolor[HTML]{EFEFEF} 
\cellcolor[HTML]{EFEFEF}                        & \cellcolor[HTML]{EFEFEF}                          & \cellcolor[HTML]{EFEFEF}                           & MotionBench      & UFVideoBench                \\
\rowcolor[HTML]{EFEFEF} 
\multirow{-2}{*}{\cellcolor[HTML]{EFEFEF}Model} & \multirow{-2}{*}{\cellcolor[HTML]{EFEFEF}MVbench} & \multirow{-2}{*}{\cellcolor[HTML]{EFEFEF}VideoMME} & (Overall / AVG.) & (PixRQA / PixHQA / PixTRQA) \\ \midrule
Qwen2.5-VL-7B                                   & 69.6                                              & \cellcolor[HTML]{FFFFFF}53.1                       & 53.0 / 48.8      & 2.54 / 3.99 / 3.78          \\
FaVChat                                         & 70.4                                              & 54.2                                               & 53.9 / 50.1      & 2.59 / 4.07 / 3.84          \\ \bottomrule
\end{tabular}
    \end{adjustbox}
\end{table*}



\section{Broader Impacts}
\label{impact}
\name{} enables precise facial video understanding by capturing subtle visual and dynamic cues such as micro-expressions, fine-grained textures, and facial motions. This capability supports human-centric applications including mental health assessment, empathetic and socially aware AI, and behavioral research, where nuanced facial signals are essential. By releasing the FaVChat-170K dataset and the data-efficient DE-GRPO framework, this work promotes reproducible research and lowers the barrier to fine-grained facial understanding, ultimately contributing to the development of assistive AI systems that augment human judgment and enhance human–machine interaction. At the same time, facial analysis technologies require careful deployment because predictions may be sensitive to demographic variation, recording conditions, and privacy constraints. We therefore encourage use in consent-based, auditable settings with human oversight, bias evaluation, and safeguards against surveillance or identity-based misuse.

\section{Limitations}
\label{limitations}

Although \name{} demonstrates strong performance in fine-grained facial video understanding, it still has several limitations. First, the current framework is built on visual-text modalities and does not explicitly incorporate audio signals. As a result, it may miss speech-related cues such as tone, prosody, pauses, and vocal affect. Second, while the proposed adapter modules help \name{} preserve the general capabilities of the underlying multimodal model and mitigate catastrophic forgetting during facial-domain specialization, the model is still primarily optimized for face-centric scenarios. For fine-grained video reasoning tasks outside the facial domain, such as detailed object manipulation, sports actions, medical procedures, or other domain-specific temporal events, \name{} may not achieve outstanding performance without dedicated training data and task-specific supervision. Future work will explore audio-visual-text modeling and broader fine-grained video adaptation with specialized data from non-facial domains.

\clearpage

\begin{table*}
    \centering
    \renewcommand{\arraystretch}{1.1}
    \caption{We randomly sample multiple qualitative comparison cases \textbf{(1/4)} and add additional color labeling to the content involving fine-grained facial features.}
    \label{more qualitive 1}
     \begin{adjustbox}{max width=0.9\linewidth} 
        \begin{tabular}{p{3cm}p{21cm}}
        \toprule[2pt]
            \multicolumn{2}{c}{\includegraphics{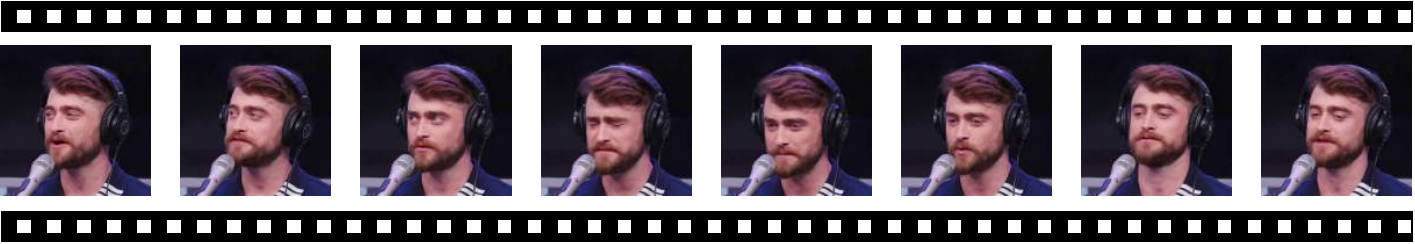}} \\
            \midrule
            \Large Question & \Large What could be the possible profession or activity of the individual in the video?\\
            \midrule
            \normalsize \textbf{FaVChat} & 
            The individual in the video presents a detailed facial profile characterized by a \textcolor{ForestGreen}{natural skin tone that hints at signs of fatigue, such as bags under the eyes}, suggesting a demanding lifestyle. His \textcolor{ForestGreen}{pointy nose adds a unique feature to his visage}, while \textcolor{ForestGreen}{his brown hair and bushy eyebrows frame his face}, contributing to a distinctive appearance. The protagonist's expressions are \textcolor{ForestGreen}{notably dynamic, ranging from a slight raise of the eyebrows and upward curve of the mouth indicating laughter}, to a \textcolor{ForestGreen}{tightening around the eyes and a slight frown that denote contemplation or concern}. These microexpressions, combined with the \textcolor{ForestGreen}{potential changes in pupil size}, reflect varying emotional states throughout the video. The \textcolor{ForestGreen}{gentle smile that forms by the final image suggests a return to a more relaxed or positive state}, highlighting the protagonist's ability to convey a spectrum of emotions. This detailed facial description, along with the professional headset and microphone, suggests that he is likely involved in a communication-intensive profession such as broadcasting, podcasting, or live streaming.\\
            \midrule
            \normalsize Qwen3-VL-8B-Face & The individual is likely engaged in radio broadcasting or podcasting. He is \textcolor{red}{wearing headphones and speaking into a professional microphone}, which are typical tools for hosting a radio show, recording a podcast, or participating in an audio interview. Given his setup, he could be a radio host, podcaster, or a guest being interviewed for an audio program.\\
            \midrule
            \normalsize GPT-4o & Based on this Video, the individual is likely engaged in a profession or activity that involves public speaking or broadcasting, such as a radio host, podcaster, or commentator. The presence of the microphone and headphones suggests that he is either recording audio content, live streaming, or participating in a broadcast where he needs to hear and communicate clearly with others or an audience. \\
            \midrule
            \normalsize Qwen3-VL-30B-A3B & The individuals in the video appear to be engaged in activities that require the use of \textcolor{red}{microphones and headphones}, which are often associated with professions or activities such as podcast recording, voice acting, radio hosting, live streaming, or music production. They may be recording podcast programs, providing audio for animations or audiobooks, serving as radio hosts or commentators, live streaming games or topic discussions on Twitch or YouTube, or participating in music production by providing vocals or voiceovers. Given their casual attire and focused expressions, it is most likely that they are engaged in podcast recording, voice acting, or live streaming. \\
                \midrule
    \end{tabular}
    \end{adjustbox}   
\end{table*}

\begin{table*}
    \centering
    \renewcommand{\arraystretch}{1.1}
    \caption{We randomly sample multiple qualitative comparison cases \textbf{(2/4)} and add additional color labeling to the content involving fine-grained facial features.}
    \label{more qualitive 2}
     \begin{adjustbox}{max width=0.98\linewidth} 
        \begin{tabular}{p{3cm}p{21cm}}
        \toprule[2pt]
            \multicolumn{2}{c}{\includegraphics{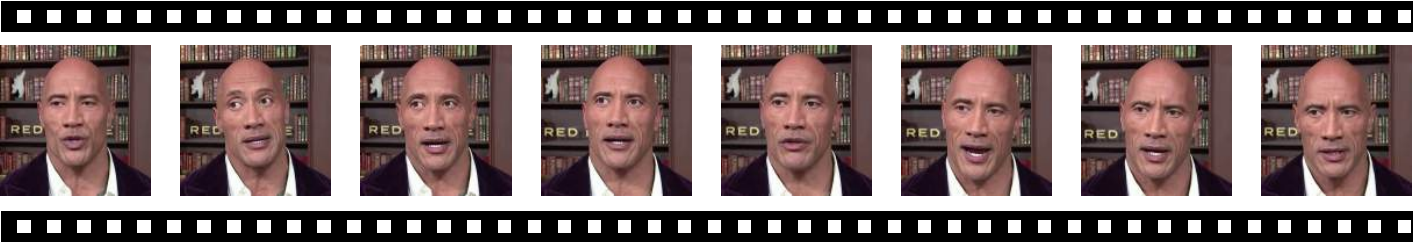}} \\
            \midrule
            \Large Question & \Large What might the protagonist be discussing? Please provide analysis basis. \\
            \midrule
            \normalsize \textbf{FaVChat} & 
            The protagonist in the video is \textcolor{ForestGreen}{a male with a bald head and a strong, mature facial structure}. His \textcolor{ForestGreen}{skin is smooth with minimal aging signs}, and his \textcolor{ForestGreen}{dark, well-defined pupils convey an intense and focused gaze}. He possesses \textcolor{ForestGreen}{a large nose with a prominent bridge and slightly rounded tip}, which, along with his \textcolor{ForestGreen}{bushy eyebrows}, forms a distinctive central feature of his face. Throughout the video, his facial expressions are \textcolor{ForestGreen}{predominantly neutral, reflecting confidence and composure}, yet he occasionally exhibits subtle microexpressions such as \textcolor{ForestGreen}{slight eyebrow movements or a gentle smirk}, indicating moments of acknowledgment or engagement in conversation. His actions, like \textcolor{ForestGreen}{looking around and nodding}, suggest curiosity and agreement, while \textcolor{ForestGreen}{talking and winking} reveal a playful and sociable side. These visual cues hint that he could be discussing a variety of subjects, potentially sharing insights or experiences related to personal growth, professional endeavors, or even leisure activities like movies or sports, all while maintaining an approachable and thoughtful demeanor.\\
            \midrule
            \normalsize Qwen3-VL-8B-Face & Judging from the background with books and the word "RED" (possibly related to a project or brand), the  protagonist might be discussing topics such as his new movie, personal brand, or a motivational speech. His expressive facial movements and formal attire suggest he is engaged in a professional or promotional conversation, probably sharing insights about his career, upcoming projects, or life philosophy.\\
            \midrule
            \normalsize GPT-4o & The protagonist appears to be engaged in a conversation or delivering a speech. His facial expressions \textcolor{red}{range from serious to smiling}, suggesting that the topic might be light-hearted or positive at times. The setting, which includes a bookshelf, indicates that the discussion could be related to literature, education, or perhaps a casual interview. The presence of the word "RED" in the background might imply that the setting is related to a brand or a specific program that uses red as a part of its visual identity. Without audio or additional context, it's challenging to determine the exact subject of the discussion, but it seems to be a friendly and possibly informative exchange. \\
            \midrule
            \normalsize Qwen3-VL-30B-A3B & The protagonist in the video frames appears to be engaged in a \textcolor{red}{serious} and possibly motivational or informative discussion. The setting, with bookshelves in the background, suggests an intellectual or professional environment, which could indicate that the topic is related to knowledge, learning, or expertise. The protagonist's attire, a suit, further supports the idea that the discussion might be formal or business-related. Given these visual cues, it is likely that the protagonist is discussing themes such as personal development, career advice, or sharing insights on a specific subject matter. The consistent and focused expression of the protagonist implies that the content of the discussion is significant and intended to engage the audience's attention. \\
                \midrule
    \end{tabular}
    \end{adjustbox}   
\end{table*}

\begin{table*}
    \centering
    \renewcommand{\arraystretch}{1.1}
    \caption{We randomly sample multiple qualitative comparison cases \textbf{(3/4)} and add additional color labeling to the content involving fine-grained facial features.}
    \label{more qualitive 3}
     \begin{adjustbox}{max width=0.98\linewidth} 
        \begin{tabular}{p{3cm}p{21cm}}
        \toprule[2pt]
            \multicolumn{2}{c}{\includegraphics{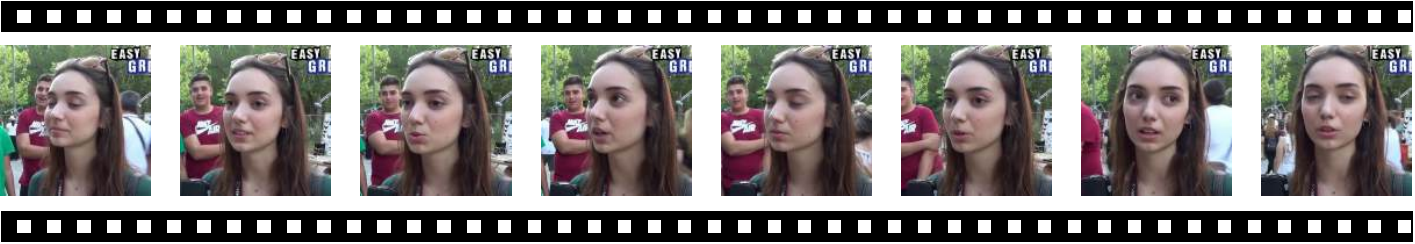}} \\
            \midrule
            \Large Question & \Large Please analyze the content of the video. \\
            \midrule
            \normalsize \textbf{FaVChat} & 
            The young woman in the video exhibits a \textcolor{ForestGreen}{clear complexion with a smooth texture}, though \textcolor{ForestGreen}{there are slight bags under her eyes that hint at potential tiredness}. Her \textcolor{ForestGreen}{long, straight brown hair} frames her face, enhancing her \textcolor{ForestGreen}{youthful appearance.} She has a \textcolor{ForestGreen}{neutral gaze}, with \textcolor{ForestGreen}{eyes that do not distinctly show their color but convey a calm and composed look}. Her \textcolor{ForestGreen}{nose is straight and well-proportioned}, contributing to a balanced facial structure. Her facial expressions are subtle, with no strong emotional indicators, which aligns with her overall \textcolor{ForestGreen}{neutral} demeanor. She \textcolor{ForestGreen}{accessorizes with sunglasses placed atop her head, adding a casual flair to her appearance}, and she \textcolor{ForestGreen}{wears small hoop earrings} that subtly complement her features. Her \textcolor{ForestGreen}{neutral expression}, combined with her physical attributes, portrays a composed and approachable individual, likely engaged in a casual yet reflective conversation amidst a lively outdoor setting, where other people and event structures can be seen in the background, suggesting a public or social gathering.\\
            \midrule
            \normalsize Qwen3-VL-8B-Face & This video features a woman being interviewed outdoors, likely in a public or recreational area with trees and other people in the background. \textcolor{red}{She has sunglasses on her head, wears a small necklace, and carries a camera strap}, suggesting she might be engaged in activities like photography or exploring.\\
            \midrule
            \normalsize GPT-4o & The video show a young woman with \textcolor{red}{long brown hair and a pair of sunglasses} resting on her head. She appears to be in an outdoor setting, possibly at a public event, as suggested by the presence of other people in the background and what looks like a stage or booth with equipment. The woman is wearing a green top and has a black bag slung over her shoulder. Her facial expressions vary across the frames, indicating she might be in the middle of a conversation or reacting to something happening off-camera. The logo in the corner suggests that the video may be from a channel or series named "EASY GRILL," which could imply an interview or street-style content format. There are no discernible texts or other specific details that provide context for the conversation or the event. \\
            \midrule
            \normalsize Qwen3-VL-30B-A3B & The video appears to show a \textcolor{red}{young woman} with \textcolor{red}{long brown hair, wearing a black top}, standing outdoors. She is \textcolor{red}{looking around} and seems to be engaged in a conversation or reacting to something off-camera. In the background, there are other people, including a man in a red shirt who is also looking around. The setting suggests a casual, possibly social environment, with trees and greenery visible, indicating it might be a park or an outdoor event. The woman's expressions change slightly across the frames, suggesting she is actively participating in the interaction. \\
                \midrule
    \end{tabular}
    \end{adjustbox}   
\end{table*}

\begin{table*}
    \centering
    \renewcommand{\arraystretch}{1.1}
    \caption{We randomly sample multiple qualitative comparison cases \textbf{(4/4)} and add additional color labeling to the content involving fine-grained facial features.}
    \label{more qualitive 4}
     \begin{adjustbox}{max width=0.98\linewidth} 
        \begin{tabular}{p{3cm}p{21cm}}
        \toprule[2pt]
            \multicolumn{2}{c}{\includegraphics{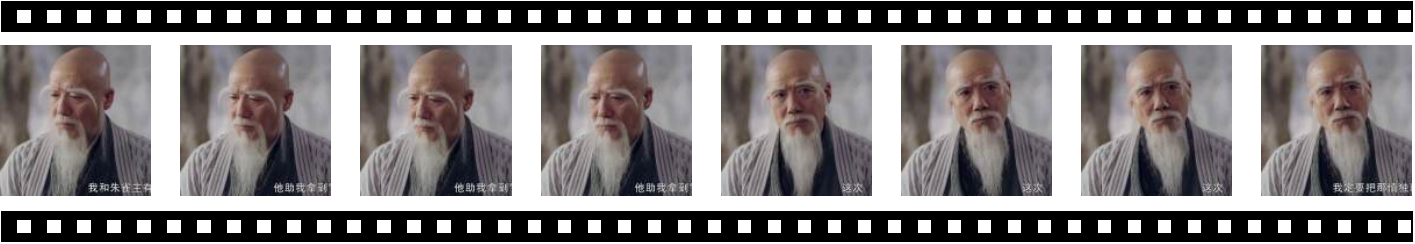}} \\
            \midrule
            \Large Question & \Large Please analyze the identity of the protagonist in the video. \\
            \midrule
            \normalsize \textbf{FaVChat} & 
            The character appears to be a figure of wisdom and experience with \textcolor{ForestGreen}{a long white beard and mustache}, as suggested by his \textcolor{ForestGreen}{oval face shape} which is traditionally associated with such qualities. His \textcolor{ForestGreen}{baldness} adds a sense of maturity and might imply a high-ranking or spiritual role. The \textcolor{ForestGreen}{bushy eyebrows and pointy nose structure} contribute to an intense and distinguished look, possibly indicating a character of authority or decisiveness. The \textcolor{ForestGreen}{goatee and mustache frame his face}, enhancing his mature appearance and potentially emphasizing his emotional state or thoughts. His \textcolor{ForestGreen}{well-kept skin} suggests self-care and maintenance, aligning with the perception of wisdom and experience. The \textcolor{ForestGreen}{neutral expression, with a calm gaze}, might suggest contemplation or introspection, inviting viewers to consider the character's internal narrative. The subtle microexpressions, \textcolor{ForestGreen}{especially around the eyes or mouth, could hint at thoughtfulness or concern}. Considering the traditional attire and the dignified presence, it is reasonable to infer that the protagonist could be a sage, a high-ranking official, or a respected elder within the narrative. \\
            \midrule
            \normalsize Qwen3-VL-8B-Face & The video shows a character who appears to be from a historical drama or film. The character is dressed in traditional Chinese attire, which suggests that the video might be set in a historical period of China. The specific identity of the protagonist cannot be determined from the video alone without additional context or information about the video's plot, setting, or production details. If you have more information or a specific question about the video, feel free to share, and I can provide further assistance.\\
            \midrule
            \normalsize GPT-4o & The protagonist is an elderly monk, identifiable by \textcolor{red}{his bald head, long white beard and eyebrows}, and traditional monk robes. His appearance and the context from the subtitles suggest he is a wise, possibly powerful figure in a martial arts or fantasy narrative, likely playing an important role in the plot, perhaps related to some secret or mission involving elements like "Shen Du" and another character referred to as "Zhu Que Lord". \\
            \midrule
            \normalsize Qwen3-VL-30B-A3B & The protagonist in the video appears to be an elderly man with \textcolor{red}{a long white beard and mustache}, wearing traditional Chinese attire. His appearance suggests that he might be a character from a historical or fantasy drama, possibly portraying a wise elder, scholar, or martial arts master. The setting and his clothing indicate a connection to ancient Chinese culture, and his demeanor conveys a sense of authority and wisdom. \\
                \midrule
    \end{tabular}
    \end{adjustbox}   
\end{table*}

\clearpage

\end{document}